\documentclass{article} 
\usepackage{iclr2027_conference,times}

\usepackage{hyperref}
\usepackage{url}
\usepackage{algorithm}
\usepackage{amsmath,amsfonts,bm}
\usepackage{algpseudocode}
\newcommand{\AlgPhase}[1]{%
    \Statex\hspace{\algorithmicindent}\textsc{#1}%
}
\usepackage{subcaption}
\usepackage{graphicx}
\usepackage{booktabs}
\usepackage{multirow}
\usepackage{setspace}
\usepackage{dashrule}
\usepackage{tikz}
\usetikzlibrary{shadows.blur}
\graphicspath{{figs/}}

\usepackage{xcolor}
\usepackage[table]{xcolor}
\definecolor{darkred}{HTML}{A31F34}
\definecolor{oursgray}{gray}{0.92}

\newcommand{\paperimage}[2][]{%
    \begin{tikzpicture}
        \node[
            inner sep=0pt,
            outer sep=0pt,
            draw=black,
            line width=0.7pt,
            blur shadow={shadow xshift=1.5pt,shadow yshift=-1.5pt,shadow blur radius=1.2pt,shadow blur steps=8,shadow opacity=35}
        ]{%
            \includegraphics[#1]{#2}%
        };
    \end{tikzpicture}%
}

\newlength{\mainimgw}
\newcommand{\methodgap}{\hspace{0\textwidth}}
\newcommand{\groupgap}{\hspace{0.015\textwidth}}

\newcommand{\methodcell}[1]{%
    \begin{minipage}[t]{\mainimgw}
        \centering\scriptsize #1
    \end{minipage}%
}
\newcommand{\imgcell}[1]{%
    \begin{minipage}[t]{\mainimgw}
        \centering
        \paperimage[width=0.99\linewidth]{#1}
    \end{minipage}%
}
\newcommand{\mainresultrow}[4]{%
    \noindent
    \methodcell{Full-Precision}
    \methodgap
    \methodcell{Quantized}
    \groupgap
    \methodcell{Bias Correction}
    \methodgap
    \methodcell{TAC-Diffusion}
    \methodgap
    \methodcell{Q-Drift}
    \methodgap
    \methodcell{\textbf{QuAKE (Ours)}}

    \vspace{2pt}

    \noindent
    \imgcell{#1/#2/fp.jpg}
    \methodgap
    \imgcell{#1/#2/quantized.jpg}
    \groupgap
    \imgcell{#1/#2/bc.jpg}
    \methodgap
    \imgcell{#1/#2/tac.jpg}
    \methodgap
    \imgcell{#1/#2/qdrift.jpg}
    \methodgap
    \imgcell{#1/#2/quake.jpg}

    \vspace{1pt}

    {\scriptsize
    Prompt: \textit{#4}\par
    }

    \vspace{3pt}

    {\centering\small #3\par}
}

\title{Quantization-Aware Kalman Estimation \\ for Diffusion Sampling}

\author{Qitan Shi, Cheng Jin, Jiawei Zhang \& Yuantao Gu\thanks{Corresponding author.} \\
Tsinghua University\\
\texttt{\{sqt24,jinc21,jiawei-z23\}@mails.tsinghua.edu.cn, gyt@tsinghua.edu.cn}
}

\iclrfinalcopy 
\begin{document}

\maketitle

\begin{abstract}
Quantization offers a practical path to deploying diffusion models with reduced memory and computation, but aggressive compression can cause quantized outputs to deviate substantially from their full-precision counterparts. Sampling-stage correction methods seek to compensate for such deviations during sampling, but existing approaches rely primarily on local information and underexploit trajectory history, limiting their ability to correct errors that propagate across timesteps. In this work, we formulate sampling with a quantized denoiser as an online estimation problem, using the history of quantized denoiser outputs to recover the underlying full-precision outputs required by the sampler.
We propose \emph{QuAKE}, a Quantization-Aware Kalman Estimator that combines a smooth trajectory prior with a conditional Gaussian observation model.
At each sampling step, QuAKE recursively updates the posterior over the output window in closed form and feeds its posterior mean to the sampler.
QuAKE is a lightweight plug-and-play corrector that requires no modification to the quantized network and naturally supports arbitrary high-order multistep ODE samplers.
Experiments across W4A4-quantized text-to-image diffusion models show that QuAKE consistently outperforms existing methods in reducing the distributional discrepancy from full-precision sampling.
\end{abstract}

\section{Introduction}
\label{sec:introduction}

Diffusion models have achieved strong performance in image synthesis, text-to-image generation, and other generative tasks~\citep{ho2020denoising,dhariwal2021diffusion,rombach2022high,saharia2022photorealistic,liu2023audioldm}.
However, deploying these models on resource-constrained hardware remains challenging because of their large memory consumption and the repeated denoiser evaluations required during iterative sampling~\citep{li2023q,he2023ptqd}.
Post-training quantization (PTQ) offers a practical path to deploying pretrained diffusion models by converting them to low precision using only a small calibration set, thereby reducing memory and inference costs without additional training~\citep{li2023q,shang2023post,li2024svdquant}.

Despite these benefits, aggressive low-bit PTQ can still substantially perturb denoiser outputs and alter the resulting sampling trajectory.
Because the denoiser output at each step determines the next sample state, quantization errors are fed back into subsequent model evaluations and can accumulate throughout sampling, shifting the generated distribution away from full-precision sampling~\citep{he2023ptqd,yao2024timestep,liu2025error}.
This has motivated sampling-stage correction methods, which use estimated quantization-error statistics to steer the quantized sampling process toward its full-precision counterpart without modifying the quantized network~\citep{he2023ptqd,yao2024timestep,zeng2025d,liu2025error,zhong2025test,ryu2026q}.

Yet existing sampling-stage correction methods are typically local in time, constructing the correction at each step primarily from the current quantized evaluation while overlooking useful temporal structure in the denoiser outputs.
Because these outputs empirically evolve smoothly along the sampling trajectory, past quantized evaluations remain informative about the underlying full-precision denoiser outputs at subsequent timesteps.

\begin{figure*}[t]
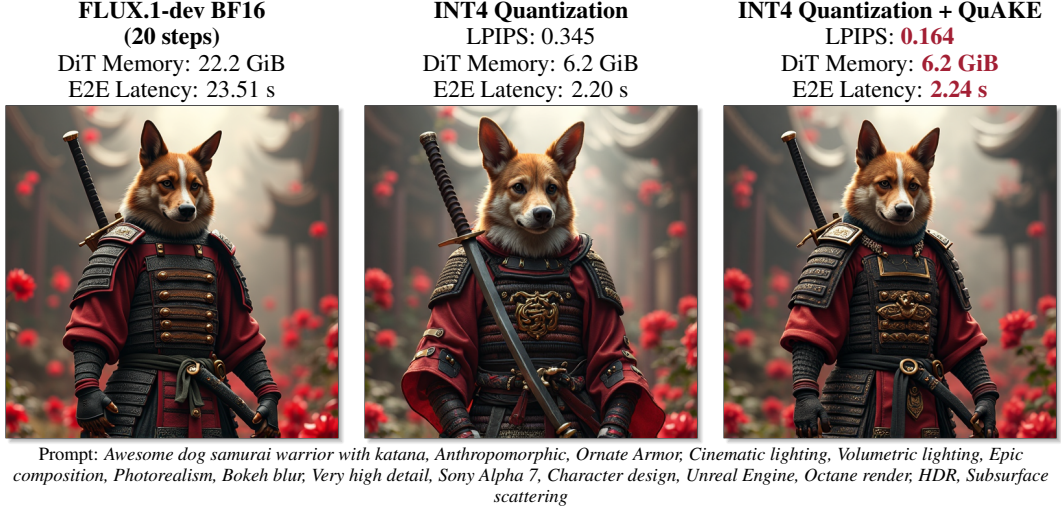

    \centering

    \begin{minipage}[t]{0.32\textwidth}
        {\centering\small
        \textbf{FLUX.1-dev BF16}\par
        \textbf{(20 steps)}\par
        DiT Memory: 22.2 GiB\par
        E2E Latency: 23.51 s\par
        }
    \end{minipage}\hfill
    \begin{minipage}[t]{0.32\textwidth}
        {\centering\small
        \textbf{INT4 Quantization}\par
        LPIPS: 0.345\par
        DiT Memory: 6.2 GiB\par
        E2E Latency: 2.20 s\par
        }
    \end{minipage}\hfill
    \begin{minipage}[t]{0.32\textwidth}
        {\centering\small
        \textbf{INT4 Quantization + QuAKE}\par
        LPIPS: \textbf{\textcolor{darkred}{0.164}}\par
        DiT Memory: \textbf{\textcolor{darkred}{6.2 GiB}}\par
        E2E Latency: \textbf{\textcolor{darkred}{2.24 s}}\par
        }
    \end{minipage}

    \vspace{0pt}

    \begin{minipage}[t]{0.32\textwidth}
        \centering
        \paperimage[width=0.98\linewidth]{teaser/fp.png}
    \end{minipage}\hfill
    \begin{minipage}[t]{0.32\textwidth}
        \centering
        \paperimage[width=0.98\linewidth]{teaser/int4.png}
    \end{minipage}\hfill
    \begin{minipage}[t]{0.32\textwidth}
        \centering
        \paperimage[width=0.98\linewidth]{teaser/quake.png}
    \end{minipage}

    {\scriptsize
    Prompt: \textit{Awesome dog samurai warrior with katana, Anthropomorphic, Ornate Armor, Cinematic lighting, Volumetric lighting, Epic composition, Photorealism, Bokeh blur, Very high detail, Sony Alpha 7, Character design, Unreal Engine, Octane render, HDR, Subsurface scattering}\par
    }

    \caption{As a lightweight plug-and-play method, QuAKE corrects quantization-induced output shifts with negligible additional overhead.}
    \label{fig:teaser}
\end{figure*}

Motivated by this observation, we formulate quantized diffusion sampling as an online state-estimation problem, where the latent state to be estimated is the full-precision denoiser-output window required by the numerical sampler, while the quantized outputs serve as noisy observations.
We propose \emph{QuAKE}, a Quantization-Aware Kalman Estimator that combines a smooth trajectory prior with a conditional Gaussian observation model.
At each sampling step, QuAKE first uses the trajectory prior to predict the current full-precision output window, then corrects this prediction using the current quantized output.
The corrected output-window estimate is then supplied to the original sampler to advance the diffusion sampling trajectory to the next sample state.
Through this recursive prediction--correction process, QuAKE continuously integrates information from the observation history into the sampling trajectory.
Because its state directly represents the output window consumed by the sampler, QuAKE naturally supports arbitrary high-order multistep ODE samplers.
As a plug-and-play sampler-side corrector, QuAKE requires no modification to the quantized network and adds only lightweight filtering operations, resulting in negligible overhead relative to a denoiser evaluation, as illustrated in Figure~\ref{fig:teaser}.

Our contributions are summarized as follows:
\begin{itemize}
    \item We formulate sampling-stage correction for quantized diffusion models as a history-conditioned online state-estimation problem, where the full-precision denoiser-output window required by the numerical sampler is sequentially inferred from quantized observations along the sampling trajectory.
    
    \item We propose QuAKE, a closed-form Kalman estimator that combines a smooth trajectory prior with a conditional Gaussian observation model to fuse information across sampling steps. As a lightweight plug-and-play corrector, QuAKE requires no modification to the quantized network and naturally supports arbitrary high-order multistep ODE samplers.
    
    \item We evaluate QuAKE across different W4A4-quantized text-to-image diffusion models, demonstrating consistent improvements over existing sampling-stage correction methods in recovering the full-precision sampling distribution.
\end{itemize}

\section{Preliminaries}
\label{sec:preliminaries}

\subsection{Diffusion ODE Sampling}

Deterministic diffusion sampling can be formulated through the probability-flow ODE~\citep{song2020score,lu2022dpm}
\begin{equation}
    \frac{\mathrm{d}x_t}{\mathrm{d}t}
    =
    \alpha_t x_t
    -
    \frac{1}{2}\beta_t^2
    \nabla_{x_t}\log p_t(x_t \mid c),
    \label{eq:diffusion_ode}
\end{equation}
where $x_t$ is the sample state at time $t$, $c$ denotes optional conditioning information, and $\alpha_t$ and $\beta_t$ are the drift and diffusion coefficients, respectively.
Sampling proceeds backward in diffusion time from a noise sample, with the score function $\nabla_{x_t}\log p_t(x_t \mid c)$ evaluated using a pretrained denoising model $f_{\theta}$.

\subsection{High-Order Multistep Samplers}

Numerical samplers solve \eqref{eq:diffusion_ode} on a decreasing time grid
$\{t_i\}_{i=0}^{L}$, where $t_0$ corresponds to the noise endpoint and $t_L$ to the data endpoint.
Let $x_i := x_{t_i}$ denote the sample state at timestep $t_i$.
The corresponding denoiser output is
\begin{equation}
    f_i
    =
    f_{\theta}(x_i,t_i,c).
\end{equation}
Modern fast samplers often employ high-order multistep ODE solvers that compute the next sample state using a window of $N$ recent denoiser outputs~\citep{lu2022dpm,zhao2023unipc,lu2025dpm}.
We denote this output window at step $i$ by
\begin{equation}
    s_i
    =
    \begin{bmatrix}
        f_i,
        f_{i-1},
        \ldots,
        f_{i-N+1}
    \end{bmatrix}^{\top},
    \label{eq:denoiser_output_window}
\end{equation}
and express the corresponding sampling update as
\begin{equation}
    x_{i+1}
    =
    \Psi_i\!\left(x_i;s_i\right),
    \label{eq:high_order_sampler}
\end{equation}
where $\Psi_i$ denotes the solver-specific update rule.

\begin{figure}[t]
    \centering
    \includegraphics[width=0.98\textwidth]{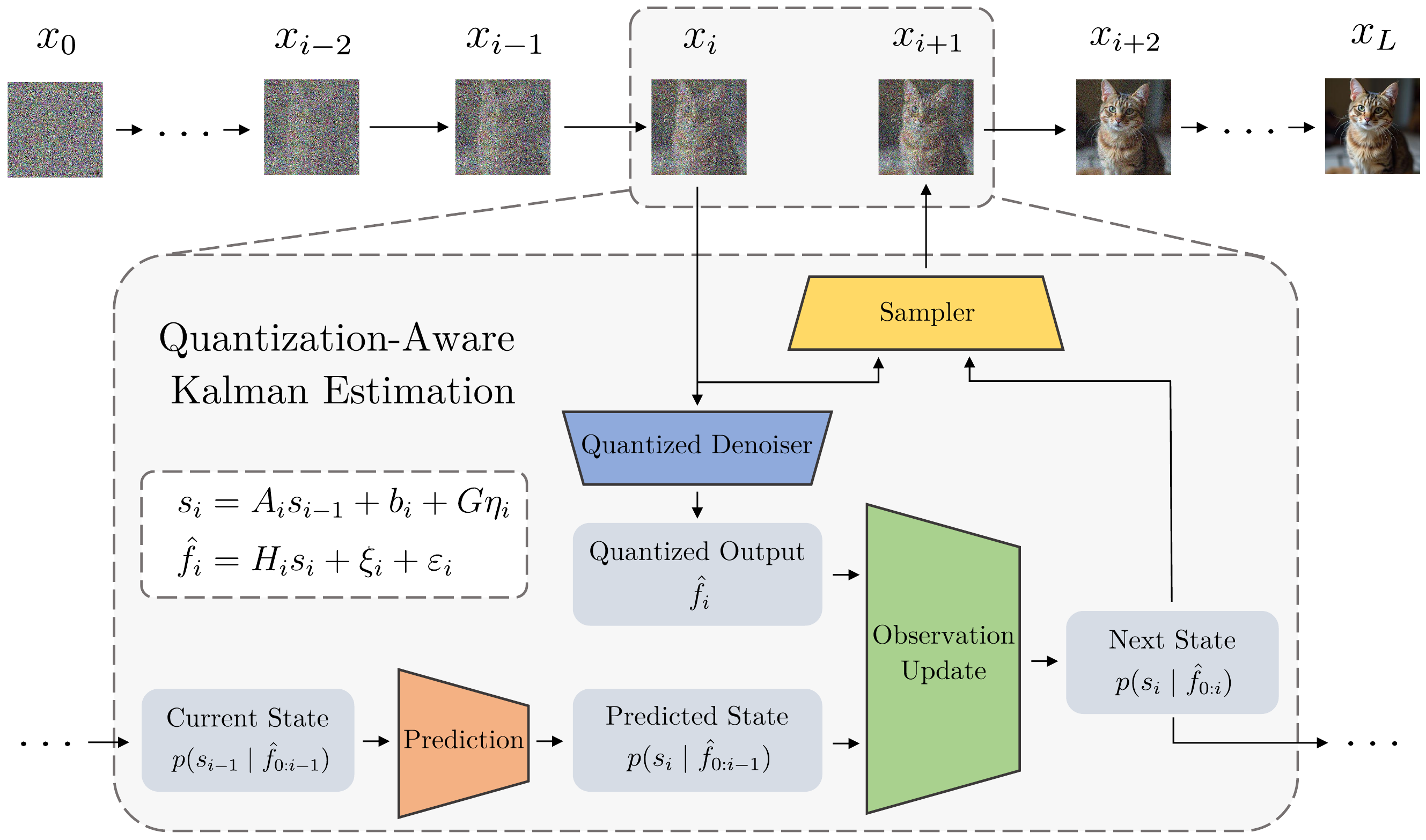}
    \caption{Overview of QuAKE. At each sampling step, QuAKE first predicts the full-precision denoiser-output window from the previous posterior, then updates the prediction using the current quantized output, and finally supplies the corrected estimate to the sampler.}
    \label{fig:pipeline}
\end{figure}

\section{Quantization-Aware Kalman Estimation}
\label{sec:quake}

In this section, we formulate QuAKE as a sequential framework for estimating the full-precision denoiser-output window from quantized outputs. We first define the estimation objective in Sec.~\ref{subsec:online_estimation}, and then establish the state-space model through the smooth trajectory prior in Sec.~\ref{subsec:smooth_trajectory_prior} and conditional Gaussian observation model in Sec.~\ref{subsec:conditional_gaussian_observation}. Finally, Sec.~\ref{subsec:recursive_posterior_estimation} derives the recursive update, and Sec.~\ref{subsec:posterior_readout} describes how the estimated outputs are supplied to the sampler.
An overview of QuAKE is illustrated in Figure~\ref{fig:pipeline}.

\subsection{Online Estimation Formulation}
\label{subsec:online_estimation}

Let $f_{\theta}$ denote the full-precision denoiser and $f_{\hat{\theta}}$ its corresponding quantized counterpart. Evaluated at the same sample state $x_i$, their outputs at step $i$ are
\begin{equation}
    f_i
    =
    f_{\theta}(x_i,t_i,c),
    \qquad
    \hat f_i
    =
    f_{\hat{\theta}}(x_i,t_i,c).
    \label{eq:full_quantized_outputs}
\end{equation}
Here, $f_i$ and $\hat f_i$ denote scalar output elements, following the element-wise treatment adopted in prior work~\citep{he2023ptqd,yao2024timestep,liu2025error,ryu2026q}.
During quantized sampling, only the quantized output $\hat f_i$ is observable, whereas the underlying full-precision output $f_i$ is unavailable.
At each sampling step $i$, the estimation target is the current full-precision denoiser-output window $s_i$ defined in \eqref{eq:denoiser_output_window}. Conditioned on all quantized outputs observed up to the current step, we formulate its estimation as
\begin{equation}
    s_i^{\star}
    =
    \arg\max_{s_i}
    p(s_i \mid \hat f_{0:i}),
    \label{eq:online_estimation_objective}
\end{equation}
where $\hat f_{0:i}:=[\hat f_0,\ldots,\hat f_i]^{\top}$ collects the complete sequence of quantized denoiser outputs available
to the estimator up to step $i$.
The estimated full-precision output window $s_i^{\star}$ is then used by the sampler to compute the next sample state 
\begin{equation}
    x_{i+1}
    =
    \Psi_i\!\left(x_i;s_i^{\star}\right).
\end{equation}
Therefore, for sampling-stage correction, given the quantized denoiser $f_{\hat{\theta}}$ and the numerical sampler updates $\{\Psi_i\}_{i=0}^{L-1}$, the task reduces to modeling and solving the estimation problem in \eqref{eq:online_estimation_objective}.

\subsection{Smooth Trajectory Prior}
\label{subsec:smooth_trajectory_prior}

Denoiser outputs along a sampling trajectory often exhibit local smoothness~\citep{liu2022pseudo,zhang2022fast,lu2022dpm,zhao2023unipc,lu2025dpm}. Motivated by this property, we assume that the current full-precision denoiser output can be linearly predicted from its recent values, with the remaining discrepancy modeled as a residual term:
\begin{equation}
    f_i
    =
    \sum_{k=1}^{N} a_{i,k} f_{i-k}
    +
    w_i,
    \qquad
    w_i \sim \mathcal{N}(\mu_i,Q_i),
    \label{eq:smooth_trajectory_model}
\end{equation}
where $\{a_{i,k}\}_{k=1}^{N}$ are predefined extrapolation coefficients, and $\mu_i$ and $Q_i$ denote the mean and variance of the prediction residual $w_i$.

Defining the zero-mean process noise
\begin{equation}
    \eta_i = w_i-\mu_i,
    \qquad
    \eta_i \sim \mathcal{N}(0,Q_i),
    \label{eq:process_noise}
\end{equation}
we can rewrite the linear extrapolation in \eqref{eq:smooth_trajectory_model} as a first-order state transition of $s_i$:
\begin{equation}
    s_i
    =
    A_i s_{i-1}
    +
    b_i
    +
    G\eta_i,
    \label{eq:state_transition}
\end{equation}
where
\begin{equation}
    A_i
    =
    \begin{bmatrix}
        a_{i,1} & a_{i,2} & \cdots & a_{i,N-1} & a_{i,N} \\
        1       & 0       & \cdots & 0         & 0         \\
        0       & 1       & \cdots & 0         & 0         \\
        \vdots  & \vdots  & \ddots & \vdots    & \vdots    \\
        0       & 0       & \cdots & 1         & 0
    \end{bmatrix},
    \qquad
    b_i=\mu_i e_1,
    \qquad
    G=e_1,
    \label{eq:transition_matrix}
\end{equation}
and $e_1=[1,0,\ldots,0]^\top$. The first row of $A_i$ extrapolates the current full-precision output from the preceding output window, while the remaining rows shift the output history forward by one step. This state transition model serves as the smooth trajectory prior for sequential estimation.

\begin{figure}[t]
    \centering
    \begin{subfigure}[t]{0.32\linewidth}
        \centering
        \includegraphics[width=\linewidth]{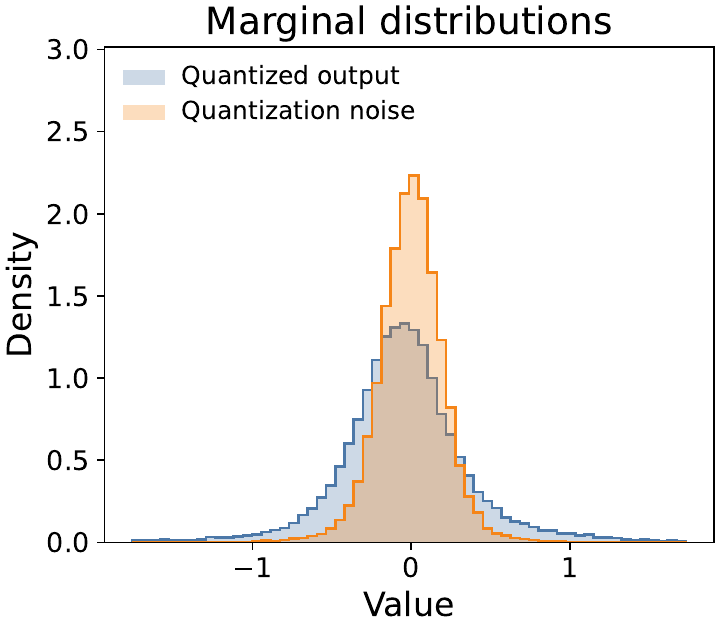}
        \caption{Marginal distributions}
        \label{fig:empirical_quant_noise_marginal}
    \end{subfigure}\hfill
    \begin{subfigure}[t]{0.32\linewidth}
        \centering
        \includegraphics[width=\linewidth]{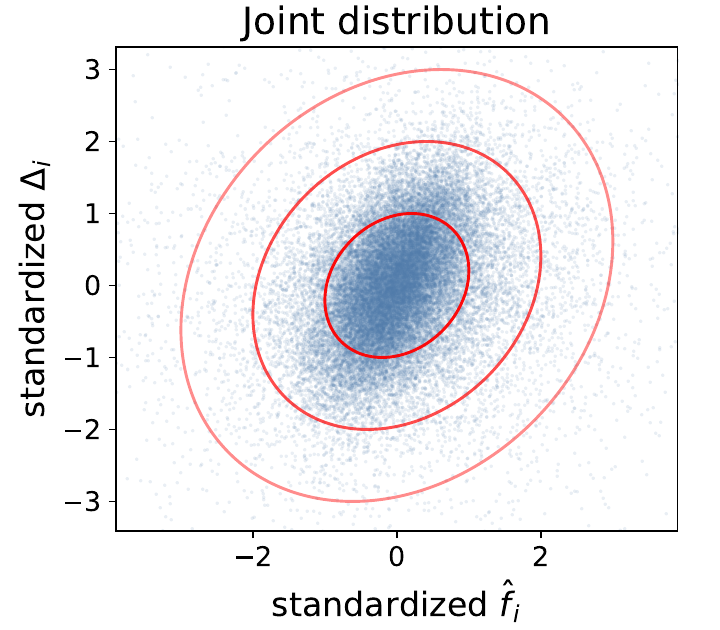}
        \caption{Joint distribution}
        \label{fig:empirical_quant_noise_joint}
    \end{subfigure}\hfill
    \begin{subfigure}[t]{0.32\linewidth}
        \centering
        \includegraphics[width=\linewidth]{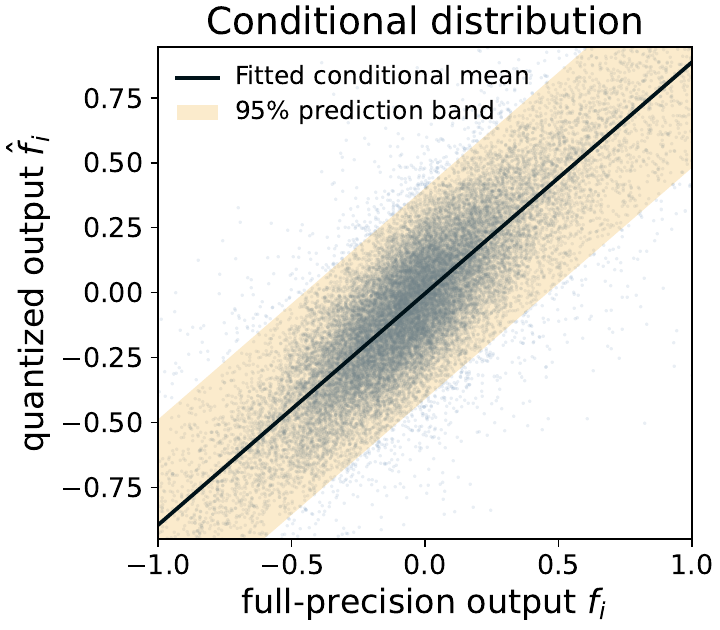}
        \caption{Conditional distribution}
        \label{fig:empirical_quant_noise_conditional}
    \end{subfigure}
    \caption{
    Empirical statistics of quantization noise.
    (a) Marginal distributions of the quantized output $\hat f_i$ and the quantization noise $\Delta_i$.
    (b) Joint distribution of $(\hat f_i,\Delta_i)$ with fitted $1\sigma$, $2\sigma$, and $3\sigma$ covariance ellipses.
    (c) Conditional distribution of $\hat f_i$ given $f_i$, with the fitted conditional mean and the $95\%$ prediction interval.
    The plots use $32{,}768$ samples from step index $5$ of Sana-0.6B and its quantized counterpart.
    }
    \label{fig:empirical_quant_noise}
\end{figure}

\subsection{Conditional Gaussian Observation}
\label{subsec:conditional_gaussian_observation}

We define the quantization noise at sampling step $i$ as $\Delta_i=\hat f_i-f_i$. Empirically, the quantized output $\hat f_i$ and the corresponding quantization noise $\Delta_i$ exhibit approximately Gaussian marginals and an elliptical joint distribution, motivating a joint Gaussian model for $(\hat f_i,\Delta_i)$~\citep{zeng2025d,ryu2026q}, as illustrated in Figure~\ref{fig:empirical_quant_noise}. Since $f_i$ is an affine transformation of $(\hat f_i,\Delta_i)$, this joint model induces the following conditional Gaussian model for $\hat f_i$ given $f_i$:
\begin{equation}
    \hat f_i \mid f_i
    \sim
    \mathcal{N}
    \left(
        \gamma_i f_i + \xi_i,
        R_i
    \right),
    \label{eq:conditional_gaussian_observation}
\end{equation}
where $\gamma_i$, $\xi_i$, and $R_i$ are timestep-dependent parameters.

Equivalently, \eqref{eq:conditional_gaussian_observation} can be expressed as the linear observation equation
\begin{equation}
    \hat f_i
    =
    \gamma_i f_i
    +
    \xi_i
    +
    \varepsilon_i,
    \qquad
    \varepsilon_i
    \sim
    \mathcal{N}(0,R_i).
    \label{eq:current_output_observation}
\end{equation}
Since the current full-precision output $f_i$ is the first component of the output-window state $s_i$, the observation equation can be written as
\begin{equation}
    \hat f_i
    =
    H_i s_i
    +
    \xi_i
    +
    \varepsilon_i,
    \qquad
    H_i
    =
    \gamma_i e_1^\top.
    \label{eq:state_observation}
\end{equation}
Thus, the current quantized denoiser output provides a noisy observation of the current full-precision output within the output-window state.

\subsection{Recursive Posterior Estimation}
\label{subsec:recursive_posterior_estimation}

Combining the state transition equation in \eqref{eq:state_transition} with the conditional Gaussian observation equation in \eqref{eq:state_observation} gives the linear Gaussian state-space system
\begin{equation}
    \begin{aligned}
        s_i
        &=
        A_i s_{i-1}
        +
        b_i
        +
        G\eta_i,
        &\qquad
        \eta_i
        &\sim
        \mathcal{N}(0,Q_i),\\
        \hat f_i
        &=
        H_i s_i
        +
        \xi_i
        +
        \varepsilon_i,
        &\qquad
        \varepsilon_i
        &\sim
        \mathcal{N}(0,R_i).
    \end{aligned}
    \label{eq:quake_state_space_model}
\end{equation}
We assume that the process and observation noises are mutually and temporally independent. Given a Gaussian initial prior for $s_0$, the posterior $p(s_i\mid\hat f_{0:i})$ remains Gaussian at every step and can be updated recursively.

\paragraph{Prediction.}
Suppose the posterior at the preceding step is
\begin{equation}
    p(s_{i-1}\mid\hat f_{0:i-1})
    =
    \mathcal{N}
    \left(
        m_{i-1|i-1},
        P_{i-1|i-1}
    \right).
    \label{eq:previous_filtering_posterior}
\end{equation}
Before observing $\hat f_i$, the predictive distribution is obtained by propagating this posterior through the state transition equation:
\begin{equation}
    p(s_i\mid\hat f_{0:i-1})
    =
    \int
    p(s_i\mid s_{i-1})
    p(s_{i-1}\mid\hat f_{0:i-1})
    \,\mathrm{d}s_{i-1}.
    \label{eq:kalman_prediction_integral}
\end{equation}
Because the transition is affine and both distributions are Gaussian, this integration gives
\begin{equation}
    p(s_i\mid\hat f_{0:i-1})
    =
    \mathcal{N}
    \left(
        m_{i|i-1},
        P_{i|i-1}
    \right),
    \label{eq:kalman_predictive_distribution}
\end{equation}
where
\begin{equation}
    m_{i|i-1}
    =
    A_i m_{i-1|i-1}
    +
    b_i,
    \qquad
    P_{i|i-1}
    =
    A_iP_{i-1|i-1}A_i^\top
    +
    GQ_iG^\top.
    \label{eq:kalman_prediction}
\end{equation}

\paragraph{Observation update.}
Once the current quantized output $\hat f_i$ is observed, Bayes' rule gives
\begin{equation}
    p(s_i\mid\hat f_{0:i})
    \propto
    p(\hat f_i\mid s_i) \;
    p(s_i\mid\hat f_{0:i-1}).
    \label{eq:kalman_bayes_update}
\end{equation}
Because the predictive distribution and observation likelihood are both Gaussian, applying Bayes' rule yields the following closed-form Kalman update. For compactness, we define the innovation $r_i$, its covariance $S_i$, and the Kalman gain $K_i$ as
\begin{equation}
    r_i
    =
    \hat f_i-H_i m_{i|i-1}-\xi_i,
    \qquad
    S_i
    =
    H_iP_{i|i-1}H_i^\top+R_i,
    \qquad
    K_i
    =
    P_{i|i-1}H_i^\top S_i^{-1}.
    \label{eq:kalman_terms}
\end{equation}
The resulting posterior is
\begin{equation}
    p(s_i\mid\hat f_{0:i})
    =
    \mathcal{N}
    \left(
        m_{i|i},
        P_{i|i}
    \right),
    \label{eq:kalman_filtering_posterior}
\end{equation}
where
\begin{equation}
    m_{i|i}
    =
    m_{i|i-1}
    +
    K_i r_i,
    \qquad
    P_{i|i}
    =
    P_{i|i-1}
    -
    K_iS_iK_i^\top.
    \label{eq:kalman_update}
\end{equation}

This recursive update aggregates the complete observation history into a compact posterior, without explicitly revisiting preceding observations. Moreover, its covariance-aware weighting adaptively balances the trajectory prediction against the current quantized observation, preserving reliable historical information while suppressing uncertain observation noise. All the required model statistics can be estimated offline through calibration. 
Detailed derivations are provided in Appendix~\ref{app:kalman_derivation}, and the calibration procedure is described in Appendix~\ref{app:calibration_statistics}.

\subsection{Posterior Readout for Sampling}
\label{subsec:posterior_readout}

Since the posterior in \eqref{eq:kalman_filtering_posterior} is Gaussian, its mean coincides with the MAP estimate in \eqref{eq:online_estimation_objective}. We therefore read out the corrected denoiser-output window as $s_i^\star=m_{i|i}$ and supply it directly to the original numerical solver:
\begin{equation}
    x_{i+1}
    =
    \Psi_i\!\left(x_i;m_{i|i}\right).
    \label{eq:quake_sampler_update}
\end{equation}
Because the posterior is defined over the complete output window, each new quantized observation updates not only the current denoiser output but also the retained historical outputs through their posterior correlations. QuAKE thus provides the high-order sampler with a temporally consistent, jointly corrected output history. Moreover, it modifies only the output window supplied to the sampler while leaving the solver-specific update rule unchanged, making it directly compatible with different high-order multistep solvers. The complete QuAKE sampling procedure is summarized in Algorithm~\ref{alg:quake}.

\begin{algorithm}[t]
    \caption{Quantized diffusion sampling with QuAKE}
    \label{alg:quake}
    \begin{algorithmic}[1]
        \Require Initial sample $x_0$, optional conditioning $c$, time grid $\{t_i\}_{i=0}^{L}$, quantized denoiser $f_{\hat{\theta}}$, solver updates $\{\Psi_i\}_{i=0}^{L-1}$, calibrated model parameters, and initial predictive statistics $m_{0|-1}$ and $P_{0|-1}$
        \Ensure Generated sample $x_L$
        \For{$i=0,\ldots,L-1$}
            \AlgPhase{\# Prediction}
            \If{$i=0$}
                \State $m_{i|i-1}\gets m_{0|-1}$,\quad
                $P_{i|i-1}\gets P_{0|-1}$
            \Else
                \State $m_{i|i-1}
                \gets
                A_i m_{i-1|i-1}+b_i$,\quad
                $P_{i|i-1}
                \gets
                A_iP_{i-1|i-1}A_i^\top+GQ_iG^\top$
            \EndIf

            \AlgPhase{\# Observation Update}
            \State $\hat f_i\gets f_{\hat{\theta}}(x_i,t_i,c)$
            \State $r_i\gets\hat f_i-H_i m_{i|i-1}-\xi_i$,\quad
            $S_i\gets H_iP_{i|i-1}H_i^\top+R_i$,\quad
            $K_i\gets P_{i|i-1}H_i^\top S_i^{-1}$
            \State $m_{i|i}\gets m_{i|i-1}+K_i r_i$,\quad
            $P_{i|i}\gets P_{i|i-1}-K_iS_iK_i^\top$

            \AlgPhase{\# Sampling}
            \State $x_{i+1}\gets\Psi_i\!\left(x_i;m_{i|i}\right)$
        \EndFor
    \end{algorithmic}
\end{algorithm}

\begin{table*}[t]
\centering
\caption{
Quantitative results on MJHQ-30K and sDCI using the DPM-Solver++ sampler. KID values are reported after multiplication by $10^3$. IR stands for ImageReward. Boldface marks the best result among sampling-stage correction methods for each model.
}
\label{tab:results_dpmsolverpp}
\scriptsize
\setlength{\tabcolsep}{3.5pt}
\renewcommand{\arraystretch}{1.18}
\begin{tabular}{cc*{10}{c}}
\toprule
\multirow{3}{*}{Model} & \multirow{3}{*}{Method}
& \multicolumn{5}{c}{MJHQ-30K} & \multicolumn{5}{c}{sDCI} \\
\cmidrule(lr){3-7}\cmidrule(lr){8-12}
& & \multicolumn{2}{c}{Similarity ($\downarrow$)} & \multicolumn{3}{c}{Quality ($\uparrow$)}
& \multicolumn{2}{c}{Similarity ($\downarrow$)} & \multicolumn{3}{c}{Quality ($\uparrow$)} \\
\cmidrule(lr){3-4}\cmidrule(lr){5-7}\cmidrule(lr){8-9}\cmidrule(lr){10-12}
& &
FID & KID & CLIPScore & CLIP-IQA & IR
& FID & KID & CLIPScore & CLIP-IQA & IR \\
\midrule
\multirow{6}{*}{\shortstack{FLUX.1-dev}} & Full-Precision & -- & -- & 0.3194 & 0.8989 & 0.9017 & -- & -- & 0.3130 & 0.8449 & 0.8589 \\
 & Quantized & 7.30 & 0.257 & 0.3179 & 0.9057 & 0.8887 & 7.22 & 0.278 & 0.3121 & 0.8540 & 0.8543 \\
\cmidrule(lr){2-12}
 & Bias Correction & 7.24 & 0.264 & 0.3183 & 0.9045 & \textbf{0.8891} & 7.21 & 0.304 & 0.3127 & 0.8513 & 0.8509 \\
 & TAC-Diffusion & 7.31 & 0.304 & 0.3179 & 0.9049 & 0.8890 & 7.21 & 0.315 & \textbf{0.3131} & \textbf{0.8544} & \textbf{0.8562} \\
 & Q-Drift & 7.16 & 0.209 & 0.3179 & \textbf{0.9051} & 0.8860 & 7.03 & 0.213 & 0.3117 & 0.8539 & 0.8501 \\
\rowcolor{oursgray}
\cellcolor{white} & QuAKE (Ours) & \textbf{7.02} & \textbf{0.158} & \textbf{0.3184} & 0.9042 & 0.8822 & \textbf{6.91} & \textbf{0.148} & 0.3127 & 0.8534 & 0.8445 \\
\midrule
\multirow{6}{*}{\shortstack{Sana-0.6B}} & Full-Precision & -- & -- & 0.3375 & 0.9003 & 1.0990 & -- & -- & 0.3310 & 0.9182 & 0.9420 \\
 & Quantized & 10.23 & 1.394 & 0.3360 & 0.8936 & 1.0481 & 10.14 & 1.465 & 0.3285 & 0.9078 & 0.8868 \\
\cmidrule(lr){2-12}
 & Bias Correction & 22.50 & 7.755 & 0.3321 & 0.8764 & 0.9522 & 23.51 & 9.246 & 0.3234 & 0.8802 & 0.7707 \\
 & TAC-Diffusion & 22.36 & 7.639 & 0.3320 & 0.8766 & 0.9576 & 23.04 & 8.970 & 0.3232 & 0.8768 & 0.7687 \\
 & Q-Drift & 9.73 & 1.072 & 0.3359 & 0.8890 & 1.0356 & 9.72 & 1.280 & 0.3285 & 0.9062 & 0.8699 \\
\rowcolor{oursgray}
\cellcolor{white} & QuAKE (Ours) & \textbf{7.90} & \textbf{0.289} & \textbf{0.3363} & \textbf{0.8951} & \textbf{1.0496} & \textbf{7.95} & \textbf{0.317} & \textbf{0.3298} & \textbf{0.9126} & \textbf{0.8753} \\
\midrule
\multirow{6}{*}{\shortstack{PixArt-$\alpha$}} & Full-Precision & -- & -- & 0.3302 & 0.8910 & 1.0004 & -- & -- & 0.3140 & 0.9141 & 0.5967 \\
 & Quantized & 15.87 & 3.177 & 0.3282 & 0.8955 & 0.9131 & 15.70 & 3.558 & 0.3118 & 0.9122 & 0.4851 \\
\cmidrule(lr){2-12}
 & Bias Correction & 28.32 & 9.511 & 0.3275 & \textbf{0.9037} & 0.8969 & 29.56 & 11.733 & 0.3146 & \textbf{0.9222} & 0.4475 \\
 & TAC-Diffusion & 28.16 & 9.573 & 0.3275 & 0.9014 & \textbf{0.9004} & 29.17 & 11.639 & \textbf{0.3149} & 0.9154 & 0.4467 \\
 & Q-Drift & 14.17 & 2.351 & 0.3278 & 0.8873 & 0.8940 & 13.93 & 2.646 & 0.3115 & 0.9032 & \textbf{0.4676} \\
\rowcolor{oursgray}
\cellcolor{white} & QuAKE (Ours) & \textbf{13.71} & \textbf{1.987} & \textbf{0.3282} & 0.8909 & 0.8934 & \textbf{13.84} & \textbf{2.401} & 0.3118 & 0.9113 & 0.4552 \\
\midrule
\multirow{6}{*}{\shortstack{PixArt-$\Sigma$}} & Full-Precision & -- & -- & 0.3318 & 0.9066 & 0.9588 & -- & -- & 0.3238 & 0.9143 & 0.8597 \\
 & Quantized & 17.22 & 4.062 & 0.3296 & 0.8978 & 0.8661 & 16.97 & 4.687 & 0.3287 & 0.8863 & 0.7943 \\
\cmidrule(lr){2-12}
 & Bias Correction & 24.38 & 7.528 & 0.3294 & 0.8891 & \textbf{0.8757} & 24.12 & 8.723 & 0.3288 & 0.8727 & 0.7755 \\
 & TAC-Diffusion & 24.07 & 7.281 & 0.3297 & 0.8889 & 0.8637 & 23.77 & 8.564 & \textbf{0.3304} & 0.8751 & 0.7808 \\
 & Q-Drift & 17.22 & 4.052 & 0.3298 & 0.8934 & 0.8546 & 16.87 & 4.592 & 0.3291 & 0.8790 & 0.7813 \\
\rowcolor{oursgray}
\cellcolor{white} & QuAKE (Ours) & \textbf{15.99} & \textbf{3.401} & \textbf{0.3299} & \textbf{0.8980} & 0.8705 & \textbf{15.50} & \textbf{3.859} & 0.3285 & \textbf{0.8863} & \textbf{0.7913} \\
\bottomrule
\end{tabular}
\end{table*}

\section{Experiments}
\label{sec:experiments}

\subsection{Experimental Setup}
\label{subsec:experimental_setup}

\paragraph{Models and datasets.}
We conduct experiments on four text-to-image diffusion models: FLUX.1-dev~\citep{flux2024}, Sana-0.6B~\citep{xie2024sana}, PixArt-$\alpha$~\citep{chen2024pixartalpha}, and PixArt-$\Sigma$~\citep{chen2024pixartsigma}. 
The full-precision models use 16-bit precision, and all models are quantized with SVDQuant~\citep{li2024svdquant} under the W4A4 setting.
Following prior work~\citep{liu2025error,ryu2026q}, we evaluate on MJHQ-30K~\citep{li2024playground} and sDCI~\citep{urbanek2024picture}, using $1{,}024$ samples from each dataset to calibrate timestep-wise statistics and a disjoint set of $5{,}000$ samples for evaluation.
For sampling, we use DPM-Solver++~\citep{lu2025dpm} in the main experiments and further evaluate UniPC~\citep{zhao2023unipc} in the supplementary experiments. Both samplers use second-order multistep updates with $20$ sampling steps.

\paragraph{Compared methods.}
We use the full-precision model and the uncorrected quantized model as two primary baselines.
We additionally compare QuAKE with representative sampling-time correction methods, including Bias Correction (PTQD~\citep{he2023ptqd} and D$^2$-DPM~\citep{zeng2025d}, which reduce to the same conditional-mean correction under the deterministic ODE sampling setting), TAC-Diffusion~\citep{yao2024timestep}, and Q-Drift~\citep{ryu2026q}.
As no official implementations of TAC-Diffusion or Q-Drift are publicly available, we reimplement both methods following their original papers.
For a fair comparison, all correction methods use the same quantized backbone, calibration and evaluation sets, sampler, initial noise samples, and number of sampling steps.

\paragraph{Metrics.}
We use FID~\citep{heusel2017gans} and KID~\citep{binkowski2018demystifying} to quantify the distributional gap between quantized generation and full-precision generation, treating the latter as the reference distribution.
We further report CLIPScore~\citep{hessel2021clipscore} to evaluate text--image semantic alignment, CLIP-IQA~\citep{wang2023exploring} to assess perceptual image quality, and ImageReward~\citep{xu2023imagereward} to measure human-preference-aligned generation quality.
Additional experimental details are provided in Appendix~\ref{app:detailed_experimental_setup}.

\begin{figure*}[t]
    \centering

    \begin{subfigure}[t]{0.24\textwidth}
        \centering
        \includegraphics[width=\linewidth]{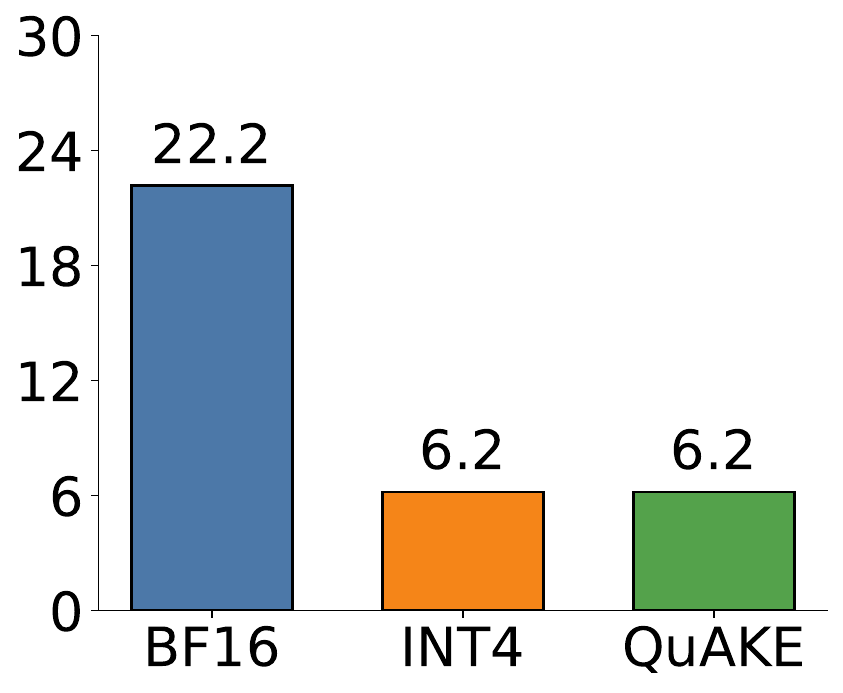}
        \caption{\footnotesize DiT Size (GiB)}
        \label{fig:dit_size}
    \end{subfigure}
    \hfill
    \begin{subfigure}[t]{0.24\textwidth}
        \centering
        \includegraphics[width=\linewidth]{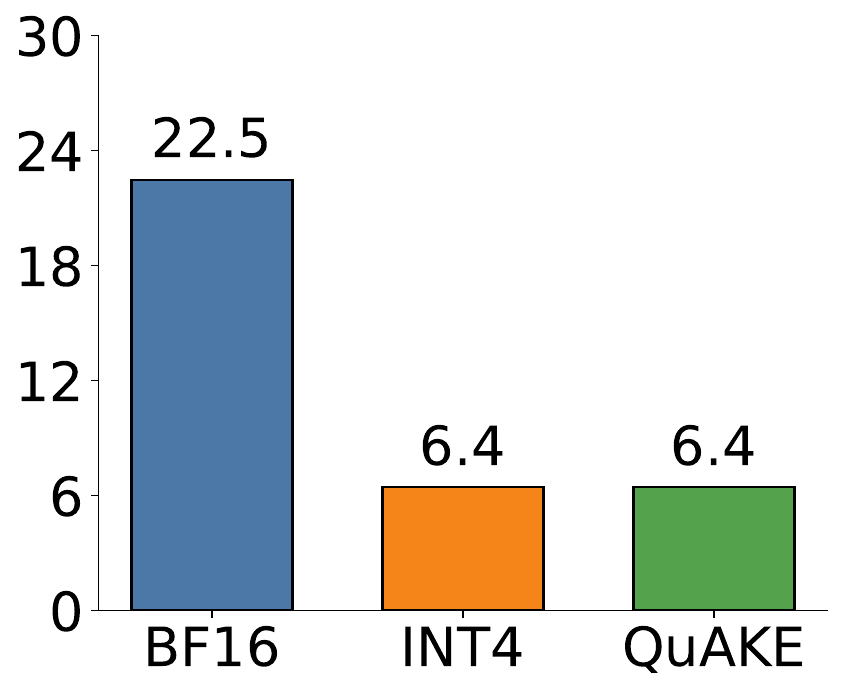}
        \caption{\footnotesize DiT Inference Memory (GiB)}
        \label{fig:dit_peak_memory}
    \end{subfigure}
    \hfill
    \begin{subfigure}[t]{0.24\textwidth}
        \centering
        \includegraphics[width=\linewidth]{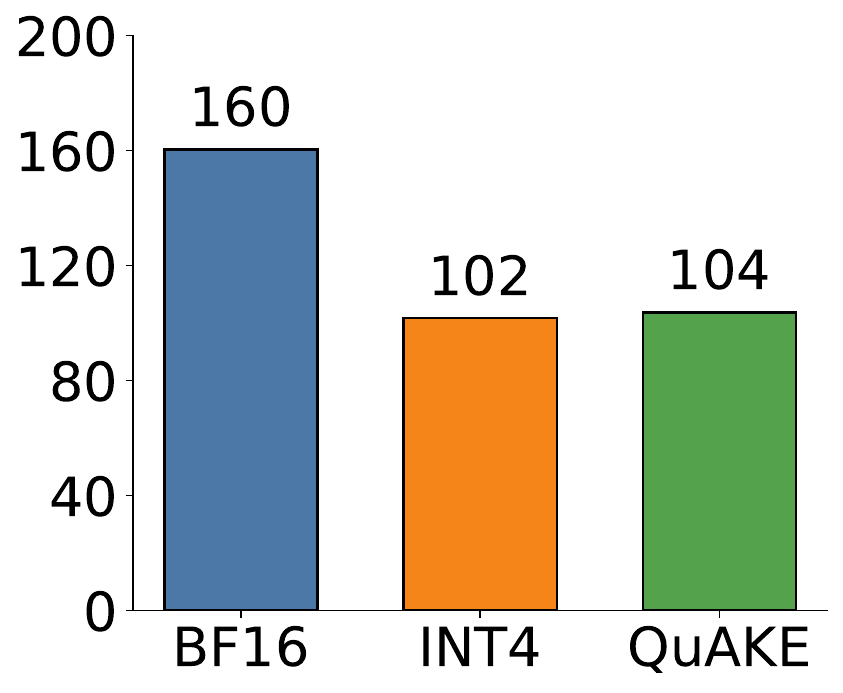}
        \caption{\footnotesize Per-Step Latency (ms)}
        \label{fig:denoise_step_latency}
    \end{subfigure}
    \hfill
    \begin{subfigure}[t]{0.24\textwidth}
        \centering
        \includegraphics[width=\linewidth]{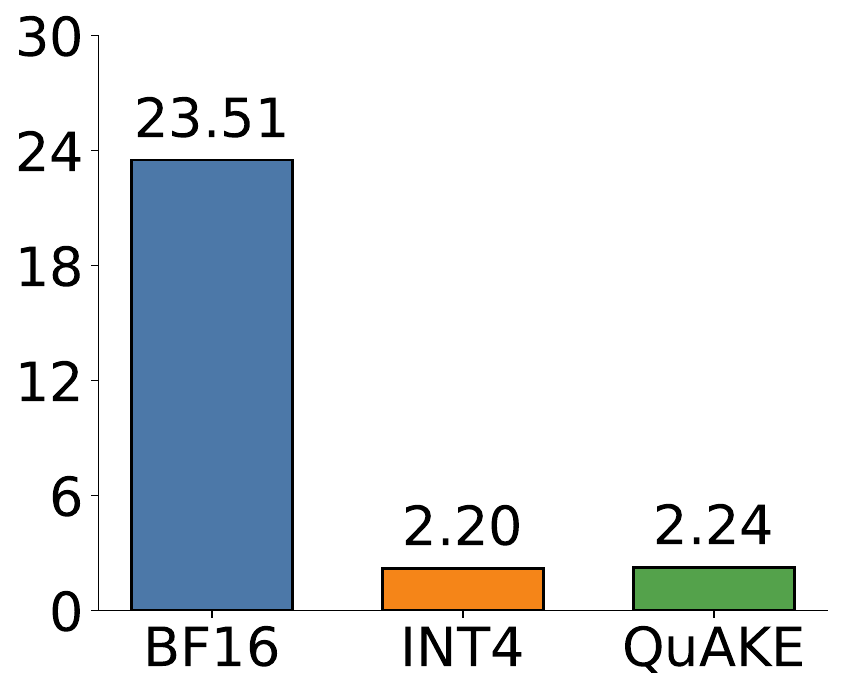}
        \caption{\footnotesize E2E Latency (s)}
        \label{fig:end_to_end_latency}
    \end{subfigure}

    \caption{
    Memory and latency measurements for generating a single image with FLUX.1-dev.
    Compared with standard quantized sampling, QuAKE introduces negligible additional memory usage and latency.
    All measurements are conducted on a single NVIDIA A100-SXM4-40GB GPU, with CPU offloading enabled for BF16 sampling.
    }
    \label{fig:histogram_comparison}
\end{figure*}

\begin{figure*}[t]
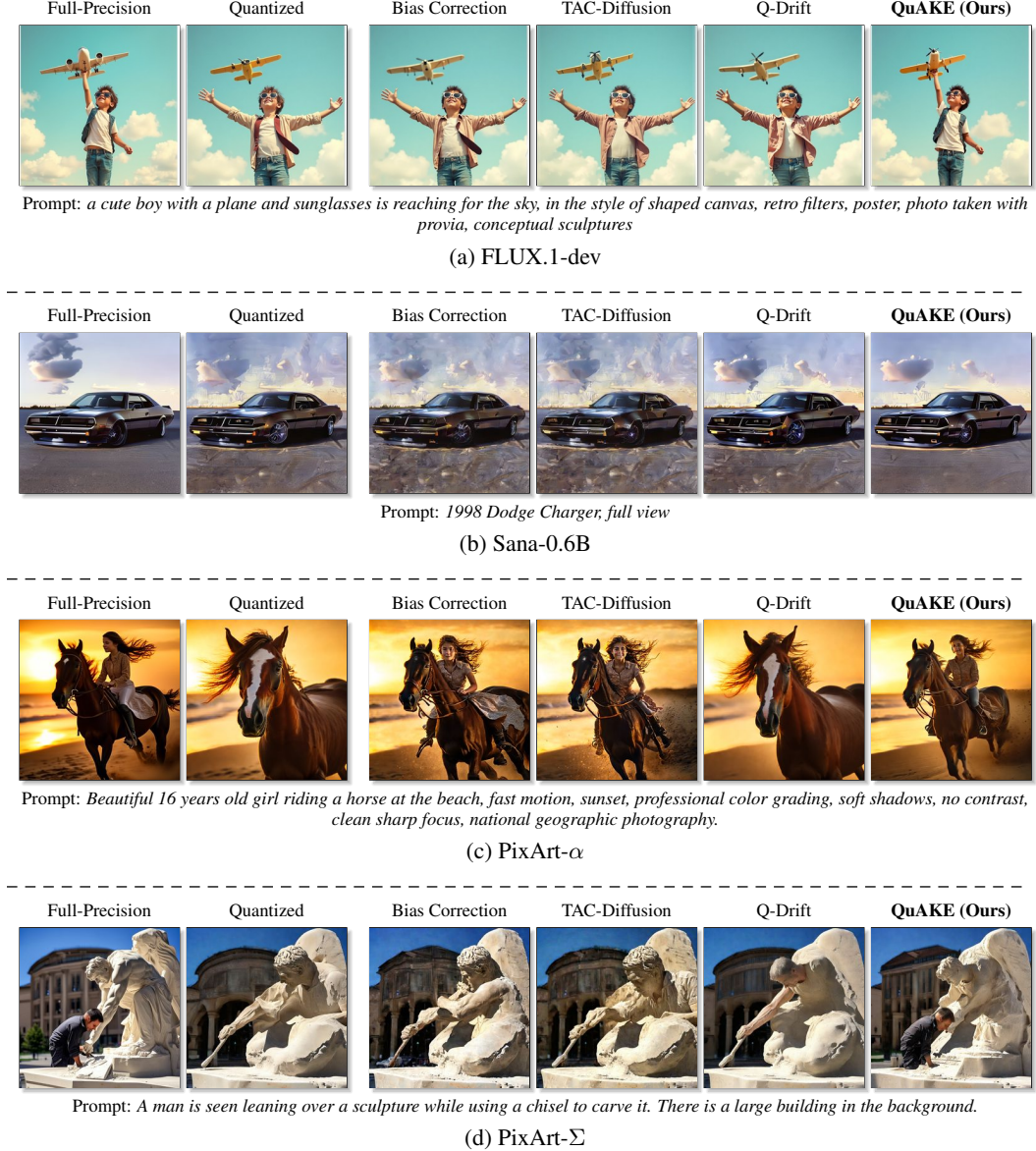

    \centering

    \mainresultrow
    {main}
    {flux}
    {(a) FLUX.1-dev}
    {a cute boy with a plane and sunglasses is reaching for the sky, in the style of shaped canvas, retro filters, poster, photo taken with provia, conceptual sculptures}

    \hdashrule{\linewidth}{0.5pt}{4pt 3pt}

    \mainresultrow
    {main}
    {sana}
    {(b) Sana-0.6B}
    {1998 Dodge Charger, full view}

    \hdashrule{\linewidth}{0.5pt}{4pt 3pt}

    \mainresultrow
    {main}
    {alpha}
    {(c) PixArt-$\alpha$}
    {Beautiful 16 years old girl riding a horse at the beach, fast motion, sunset, professional color grading, soft shadows, no contrast, clean sharp focus, national geographic photography.}

    \hdashrule{\linewidth}{0.5pt}{4pt 3pt}

    \mainresultrow
    {main}
    {sigma}
    {(d) PixArt-$\Sigma$}
    {A man is seen leaning over a sculpture while using a chisel to carve it. There is a large building in the background.}

    \caption{
    Qualitative comparison across different text-to-image diffusion models using DPM-Solver++.
    QuAKE effectively compensates for the generation deviations introduced by quantization, producing results closer to the full-precision outputs than existing sampling-stage correction methods.}
    
    \label{fig:main_results}
\end{figure*}

\subsection{Main Results}
\label{subsec:main_results}

Table~\ref{tab:results_dpmsolverpp} reports the main quantitative results with DPM-Solver++ across four text-to-image diffusion models and two evaluation datasets. QuAKE consistently achieves the lowest FID and KID among all sampling-stage correction methods for every model--dataset combination, demonstrating its effectiveness in recovering the output distribution of the full-precision model. Meanwhile, QuAKE remains competitive across semantic alignment, perceptual quality, and human preference. These results indicate that \textbf{QuAKE substantially reduces the distributional discrepancy induced by quantization} without compromising overall generation quality, thereby better preserving the behavior of the full-precision model. Appendix~\ref{app:additional_results} further evaluates QuAKE with the UniPC sampler and observes a consistent overall trend, supporting its generalization across samplers and its applicability to different high-order multistep ODE solvers without solver-specific redesign.

The results also highlight the importance of maintaining temporal consistency when correcting quantized outputs for high-order multistep samplers. Bias Correction and TAC-Diffusion primarily correct the current denoiser evaluation, without explicitly accounting for its consistency with the historical outputs retained by the sampler, as detailed in Appendix~\ref{app:sampling_stage_correction}. Since high-order multistep updates jointly rely on the current and previous denoiser outputs, independently correcting the current output can introduce inconsistencies within the output history, which may in turn degrade the resulting sampling trajectory. This behavior is reflected by the increased FID and KID of these methods on several models. In contrast, QuAKE treats the complete output window as the estimation target and recursively combines a smooth trajectory prior with successive quantized observations, producing a temporally consistent corrected history. This enables our method to consistently reduce both FID and KID over the uncorrected quantized baseline across all model--dataset combinations.

Figure~\ref{fig:histogram_comparison} evaluates the memory and latency overhead introduced by QuAKE. Compared with standard quantized sampling, the calibrated statistics required by our method occupy only 0.25~MiB of additional memory for FLUX.1-dev, while its additional latency is negligible compared with the cost of a denoising network evaluation. This confirms that our method can serve as a lightweight plug-and-play correction module, providing substantial improvements in distributional recovery while largely preserving the memory and computational advantages of low-precision inference.

Figure~\ref{fig:main_results} presents qualitative comparisons across different text-to-image diffusion models. Quantization can noticeably perturb the generated outputs from their full-precision counterparts, leading to potential deviations in visual content and structure. In contrast, QuAKE effectively compensates for these quantization-induced deviations and produces images that more closely resemble the corresponding full-precision outputs. Compared with existing sampling-stage correction methods, our method better preserves the visual characteristics of the full-precision generation, consistent with the reduced distributional discrepancy observed in FID and KID.

\section{Conclusion}
\label{sec:conclusion}

In this work, we propose QuAKE, a Quantization-Aware Kalman Estimator for sampling-stage correction of quantized diffusion models. QuAKE formulates quantized diffusion sampling as an online estimation problem, enabling correction to exploit the temporal structure and historical information along the sampling trajectory. Building on this formulation, QuAKE combines a smooth trajectory prior with a conditional Gaussian observation model to recursively refine quantized outputs, while remaining lightweight, plug-and-play, and compatible with different high-order multistep ODE samplers. Experiments across multiple W4A4-quantized text-to-image diffusion models show that QuAKE consistently reduces the distributional discrepancy from full-precision generation compared with existing correction methods, with negligible memory and latency overhead.

\bibliography{iclr2027_conference}
\bibliographystyle{iclr2027_conference}

\clearpage
\appendix

\section{Related Work}
\label{app:related_work}

This appendix discusses related work on high-order ODE samplers and sampling-stage correction methods for quantized diffusion models.

\subsection{High-Order Multistep ODE Samplers for Diffusion Models}
\label{app:high_order_multistep_samplers}

High-order multistep ODE samplers reuse recent denoiser evaluations to construct higher-order approximations to the diffusion ODE solution, thereby reducing discretization error and enabling accurate sampling with fewer steps~\citep{liu2022pseudo,zhang2022fast,lu2022dpm,zhao2023unipc,lu2025dpm}.

Take DPM-Solver++~\citep{lu2025dpm} as an example. Under the data-prediction parameterization, the reverse-time ODE can be written as
\begin{equation}
    \mathrm{d}x_t
    =
    x_t\,\mathrm{d}\log\sigma_t
    +
    \sigma_t e^{\lambda_t}
    f_{\theta}(x_t,t,c)\,\mathrm{d}\lambda_t,
    \label{eq:app_data_prediction_ode}
\end{equation}
where $\sigma_t$ and $\lambda_t$ denote the marginal noise scale and log-SNR coordinate, respectively. Define
\begin{equation}
    \sigma_i:=\sigma_{t_i},\qquad
    \lambda_i:=\lambda_{t_i},\qquad
    f(\lambda):=f_{\theta}\!\left(x_{t(\lambda)},t(\lambda),c\right),\qquad
    f_i:=f(\lambda_i).
    \label{eq:app_dpmsolverpp_notation}
\end{equation}
Integrating \eqref{eq:app_data_prediction_ode} from $t_i$ to $t_{i+1}$ yields
\begin{equation}
    x_{i+1}
    =
    \frac{\sigma_{i+1}}{\sigma_i}x_i
    +
    \sigma_{i+1}
    \int_{\lambda_i}^{\lambda_{i+1}}
    e^\lambda f(\lambda)\,\mathrm{d}\lambda.
    \label{eq:app_dpmsolverpp_exact}
\end{equation}

To construct an $N$-th order update, DPM-Solver++ applies an $(N-1)$-th order Taylor expansion around $\lambda_i$:
\begin{equation}
    f(\lambda)=\sum_{n=0}^{N-1}\frac{(\lambda-\lambda_i)^n}{n!}f_i^{(n)}+\mathcal{O}\!\left((\lambda-\lambda_i)^N\right),
    \label{eq:app_dpmsolverpp_taylor}
\end{equation}
where
\begin{equation}
    f_i^{(n)}:=\left.\frac{\mathrm{d}^n f(\lambda)}{\mathrm{d}\lambda^n}\right|_{\lambda=\lambda_i}
    \label{eq:app_dpmsolverpp_derivative}
\end{equation}
denotes the $n$-th total derivative of the data prediction along the sampling trajectory. Substituting \eqref{eq:app_dpmsolverpp_taylor} into \eqref{eq:app_dpmsolverpp_exact} gives
\begin{equation}
    x_{i+1}=\frac{\sigma_{i+1}}{\sigma_i}x_i+\sigma_{i+1}\sum_{n=0}^{N-1}f_i^{(n)}\int_{\lambda_i}^{\lambda_{i+1}}e^\lambda\frac{(\lambda-\lambda_i)^n}{n!}\,\mathrm{d}\lambda+\mathcal{O}\!\left((\lambda_{i+1}-\lambda_i)^{N+1}\right).
    \label{eq:app_dpmsolverpp_high_order}
\end{equation}
The exponentially weighted integrals can be evaluated analytically, while the derivatives are approximated by finite differences of the $N$ recent denoiser evaluations. Absorbing the resulting integration and finite-difference coefficients into $\{\omega_{i,j}\}_{j=0}^{N-1}$ yields the multistep DPM-Solver++ update
\begin{equation}
    x_{i+1}=\frac{\sigma_{i+1}}{\sigma_i}x_i+\sum_{j=0}^{N-1}\omega_{i,j}f_{i-j}.
    \label{eq:app_dpmsolverpp_multistep}
\end{equation}

More generally, other high-order multistep ODE samplers can also be formulated in this form using the denoiser-output window $s_i=[f_i,f_{i-1},\ldots,f_{i-N+1}]^\top$. We uniformly abstract them throughout this work as
\begin{equation}
    x_{i+1}
    =
    \Psi_i\!\left(x_i;s_i\right).
    \label{eq:app_high_order_multistep_abstract}
\end{equation}

\paragraph{Connection to QuAKE.}
The connection to QuAKE is twofold. First, because high-order multistep samplers require the entire output window $s_i$ to advance the sampling trajectory, we take $s_i$ rather than only the current denoiser output $f_i$ as the estimation target. Second, their Taylor-based constructions implicitly exploit the local smoothness of denoiser outputs along the sampling trajectory, which motivates the smooth trajectory prior introduced in Section~\ref{subsec:smooth_trajectory_prior}.

\subsection{Sampling-Stage Correction for Quantized Diffusion Models}
\label{app:sampling_stage_correction}

Sampling-stage correction methods compensate for quantization-induced deviations by modifying the sampling process while leaving the quantized denoiser unchanged. Representative methods are described below using the formulation adopted in this work.

\paragraph{Bias correction.}
PTQD~\citep{he2023ptqd} and D$^2$-DPM~\citep{zeng2025d} model the conditional statistics of quantization noise and use the estimated conditional mean to correct the quantized denoiser output. Their complete correction rules are designed for stochastic DDPM-style sampling and additionally adjust the stochastic noise term in the sampling equation. Under the deterministic ODE sampling setting considered in this work, the conditional-mean correction can be written as
\begin{equation}
    \tilde f_i
    =
    \hat f_i
    -
    \mathbb{E}
    [
        \Delta_i
        \mid
        \hat f_i
    ].
    \label{eq:app_bias_correction}
\end{equation}
The corrected output $\tilde f_i$ is stored in the sampler output cache in place of the raw quantized output.

\paragraph{TAC-Diffusion.}
TAC-Diffusion~\citep{yao2024timestep} decomposes sampling-stage correction into Noise Estimation Reconstruction (NER) and Input Bias Correction (IBC). NER rescales the quantized denoiser output to reduce its discrepancy from the full-precision output, whereas IBC shifts the denoiser input to compensate for accumulated input bias. Under the present formulation, the resulting corrected denoiser evaluation can be written as
\begin{equation}
    \tilde f_i
    =
    \kappa_i
    f_{\hat{\theta}}
    \left(
        x_i-B_i,
        t_i,
        c
    \right),
    \label{eq:app_tac_correction}
\end{equation}
where $\kappa_i$ is the timestep-dependent reconstruction coefficient and $B_i$ is the timestep-dependent input bias. The corrected evaluation $\tilde f_i$ is then supplied to the sampler.

\paragraph{Q-Drift.}
Q-Drift~\citep{ryu2026q} treats quantization error as an implicit stochastic perturbation of the sampling dynamics and compensates for the resulting distribution shift through a marginal-preserving drift rescaling. For the $N$-output window used by a multistep sampler, the correction can be written as
\begin{equation}
    \begin{bmatrix}
        \tilde f_i,
        \tilde f_{i-1},
        \ldots,
        \tilde f_{i-N+1}
    \end{bmatrix}^\top
    =
    (1+\rho_i)
    \begin{bmatrix}
        \hat f_i,
        \hat f_{i-1},
        \ldots,
        \hat f_{i-N+1}
    \end{bmatrix}^\top,
    \label{eq:app_qdrift_correction}
\end{equation}
where $\rho_i$ is a timestep-dependent drift-rescaling coefficient determined by the conditional variance statistic and the coefficients of the numerical sampler. The rescaled outputs are subsequently used in the sampler output cache. Because $\rho_i$ depends on the discretization coefficients of the sampling update, applying Q-Drift to a different sampler requires a corresponding solver-specific derivation.

\paragraph{DDIM-specific corrections.}
Several other sampling-stage correction methods derive their correction rules directly from the first-order DDIM recursion. TCEC~\citep{liu2025error} and IEC~\citep{zhong2025test} estimate and compensate for the accumulated sample-state error induced by quantization, while StepbaQ~\citep{chen2024stepbaq} adjusts the sampling steps according to a DDIM-based quantization-error analysis. Since these correction rules explicitly depend on the coefficients and error-propagation structure of the DDIM update, they do not directly provide correction mechanisms for general high-order multistep ODE samplers.

\section{Detailed Derivations}
\label{app:kalman_derivation}

This appendix first derives the conditional Gaussian observation model in Sec.~\ref{subsec:conditional_gaussian_observation}, followed by the recursive posterior update in Sec.~\ref{subsec:recursive_posterior_estimation}.
In addition, we prove that our method remains LMMSE-optimal when the process and observation noises are not strictly Gaussian.

\subsection{Conditional Gaussian Observation}

We first derive how the joint Gaussian model of the quantized output and quantization noise induces the conditional Gaussian observation model in \eqref{eq:conditional_gaussian_observation}. Recall that the quantization noise is defined as $\Delta_i=\hat f_i-f_i$. We model the quantized output and quantization noise as jointly Gaussian:
\begin{equation}
    \begin{bmatrix}
        \hat f_i\\
        \Delta_i
    \end{bmatrix}
    \sim
    \mathcal{N}
    \left(
        \begin{bmatrix}
            \mathbb{E}[\hat f_i]\\
            \mathbb{E}[\Delta_i]
        \end{bmatrix},
        \begin{bmatrix}
            \operatorname{Var}(\hat f_i)
            &
            \operatorname{Cov}(\hat f_i,\Delta_i)\\
            \operatorname{Cov}(\hat f_i,\Delta_i)
            &
            \operatorname{Var}(\Delta_i)
        \end{bmatrix}
    \right).
    \label{eq:app_joint_quantization_model}
\end{equation}
Since $f_i=\hat f_i-\Delta_i$, the pair $(f_i,\hat f_i)$ is obtained through the linear transformation
\begin{equation}
    \begin{bmatrix}
        f_i\\
        \hat f_i
    \end{bmatrix}
    =
    \begin{bmatrix}
        1 & -1\\
        1 & 0
    \end{bmatrix}
    \begin{bmatrix}
        \hat f_i\\
        \Delta_i
    \end{bmatrix},
    \label{eq:app_full_quantized_transform}
\end{equation}
and is therefore also jointly Gaussian. The corresponding moments are
\begin{align}
    \mathbb{E}[f_i]
    &=
    \mathbb{E}[\hat f_i]
    -
    \mathbb{E}[\Delta_i],
    \label{eq:app_full_precision_mean}\\
    \operatorname{Var}(f_i)
    &=
    \operatorname{Var}(\hat f_i)
    +
    \operatorname{Var}(\Delta_i)
    -
    2\operatorname{Cov}(\hat f_i,\Delta_i),
    \label{eq:app_full_precision_variance}\\
    \operatorname{Cov}(f_i,\hat f_i)
    &=
    \operatorname{Var}(\hat f_i)
    -
    \operatorname{Cov}(\hat f_i,\Delta_i).
    \label{eq:app_full_quantized_covariance}
\end{align}
Applying the conditional Gaussian formula gives
\begin{equation}
    \hat f_i\mid f_i
    \sim
    \mathcal{N}
    \left(
        \mathbb{E}[\hat f_i]
        +
        \frac{\operatorname{Cov}(f_i,\hat f_i)}
             {\operatorname{Var}(f_i)}
        \bigl(f_i-\mathbb{E}[f_i]\bigr),
        \operatorname{Var}(\hat f_i)
        -
        \frac{\operatorname{Cov}(f_i,\hat f_i)^2}
             {\operatorname{Var}(f_i)}
    \right).
    \label{eq:app_conditional_observation_distribution}
\end{equation}
Defining
\begin{equation}
    \gamma_i
    =
    \frac{\operatorname{Cov}(f_i,\hat f_i)}
         {\operatorname{Var}(f_i)},
    \qquad
    \xi_i
    =
    \mathbb{E}[\hat f_i]
    -
    \gamma_i\mathbb{E}[f_i],
    \qquad
    R_i
    =
    \operatorname{Var}(\hat f_i)
    -
    \frac{\operatorname{Cov}(f_i,\hat f_i)^2}
         {\operatorname{Var}(f_i)},
    \label{eq:app_observation_parameters}
\end{equation}
we obtain
\begin{equation}
    \hat f_i\mid f_i
    \sim
    \mathcal{N}
    \left(
        \gamma_i f_i+\xi_i,
        R_i
    \right).
    \label{eq:app_conditional_gaussian_observation}
\end{equation}

\subsection{Recursive Posterior Estimation}

This subsection provides a detailed derivation of the recursive posterior update. Recall the linear Gaussian state-space model
\begin{equation}
    \begin{aligned}
        s_i
        &=
        A_i s_{i-1}
        +
        b_i
        +
        G\eta_i,
        &\qquad
        \eta_i
        &\sim
        \mathcal{N}(0,Q_i),\\
        \hat f_i
        &=
        H_i s_i
        +
        \xi_i
        +
        \varepsilon_i,
        &\qquad
        \varepsilon_i
        &\sim
        \mathcal{N}(0,R_i).
    \end{aligned}
    \label{eq:app_state_space_model}
\end{equation}
The process and observation noises are assumed to be mutually and temporally independent. In particular, $\eta_i$ is independent of $s_{i-1}$ and the preceding observations, while $\varepsilon_i$ is independent of $s_i$ and the preceding observations. The recursion starts from the Gaussian prior
\begin{equation}
    s_0
    \sim
    \mathcal{N}(m_{0|-1},P_{0|-1}),
    \label{eq:app_initial_prior}
\end{equation}
which serves directly as the predictive distribution at $i=0$.

\paragraph{Prediction.}
For $i\geq 1$, suppose the posterior at the preceding step is
\begin{equation}
    s_{i-1}\mid\hat f_{0:i-1}
    \sim
    \mathcal{N}
    \left(
        m_{i-1|i-1},
        P_{i-1|i-1}
    \right).
    \label{eq:app_previous_posterior}
\end{equation}
The predictive distribution is obtained by propagating this posterior through the state transition:
\begin{equation}
    p(s_i\mid\hat f_{0:i-1})
    =
    \int
    p(s_i\mid s_{i-1})
    p(s_{i-1}\mid\hat f_{0:i-1})
    \,\mathrm{d}s_{i-1}.
    \label{eq:app_prediction_integral}
\end{equation}
Because the transition is affine Gaussian, the predictive distribution remains Gaussian. Its mean is
\begin{align}
    m_{i|i-1}
    &=
    \mathbb{E}[s_i\mid\hat f_{0:i-1}]
    \nonumber\\
    &=
    \mathbb{E}
    \bigl[
        A_i s_{i-1}
        +
        b_i
        +
        G\eta_i
        \mid
        \hat f_{0:i-1}
    \bigr]
    \nonumber\\
    &=
    A_i m_{i-1|i-1}
    +
    b_i,
    \label{eq:app_prediction_mean}
\end{align}
where $\mathbb{E}[\eta_i]=0$. Its covariance is
\begin{align}
    P_{i|i-1}
    &=
    \operatorname{Cov}(s_i\mid\hat f_{0:i-1})
    \nonumber\\
    &=
    \operatorname{Cov}
    \bigl(
        A_i s_{i-1}
        +
        G\eta_i
        \mid
        \hat f_{0:i-1}
    \bigr)
    \nonumber\\
    &=
    A_iP_{i-1|i-1}A_i^\top
    +
    GQ_iG^\top,
    \label{eq:app_prediction_covariance}
\end{align}
where the cross-covariance between $s_{i-1}$ and $\eta_i$ vanishes by independence. Therefore,
\begin{equation}
    s_i\mid\hat f_{0:i-1}
    \sim
    \mathcal{N}
    \left(
        m_{i|i-1},
        P_{i|i-1}
    \right).
    \label{eq:app_predictive_distribution}
\end{equation}

\paragraph{Observation update.}
After observing $\hat f_i$, Bayes' rule gives
\begin{equation}
    p(s_i\mid\hat f_{0:i})
    =
    \frac{
        p(\hat f_i\mid s_i)
        p(s_i\mid\hat f_{0:i-1})
    }{
        p(\hat f_i\mid\hat f_{0:i-1})
    }.
    \label{eq:app_bayes_update}
\end{equation}
From the observation equation, the predictive observation mean is
\begin{align}
    \mathbb{E}[\hat f_i\mid\hat f_{0:i-1}]
    &=
    \mathbb{E}
    \bigl[
        H_i s_i
        +
        \xi_i
        +
        \varepsilon_i
        \mid
        \hat f_{0:i-1}
    \bigr]
    \nonumber\\
    &=
    H_i m_{i|i-1}
    +
    \xi_i.
    \label{eq:app_predictive_observation_mean}
\end{align}
Its covariance is
\begin{align}
    \operatorname{Cov}(\hat f_i\mid\hat f_{0:i-1})
    &=
    \operatorname{Cov}
    \bigl(
        H_i s_i+\varepsilon_i
        \mid
        \hat f_{0:i-1}
    \bigr)
    \nonumber\\
    &=
    H_iP_{i|i-1}H_i^\top
    +
    R_i,
    \label{eq:app_predictive_observation_covariance}
\end{align}
and the state--observation cross-covariance is
\begin{align}
    \operatorname{Cov}(s_i,\hat f_i\mid\hat f_{0:i-1})
    &=
    \operatorname{Cov}
    \bigl(
        s_i,H_i s_i+\varepsilon_i
        \mid
        \hat f_{0:i-1}
    \bigr)
    \nonumber\\
    &=
    P_{i|i-1}H_i^\top.
    \label{eq:app_state_observation_covariance}
\end{align}

Define the innovation and its covariance as
\begin{equation}
    r_i
    =
    \hat f_i
    -
    H_i m_{i|i-1}
    -
    \xi_i,
    \qquad
    S_i
    =
    H_iP_{i|i-1}H_i^\top
    +
    R_i,
    \label{eq:app_innovation}
\end{equation}
and define the Kalman gain as
\begin{equation}
    K_i
    =
    P_{i|i-1}H_i^\top S_i^{-1}.
    \label{eq:app_kalman_gain}
\end{equation}
Conditioned on the preceding observations, $s_i$ and $\hat f_i$ are jointly Gaussian:
\begin{equation}
    \begin{bmatrix}
        s_i\\
        \hat f_i
    \end{bmatrix}
    \Bigm|
    \hat f_{0:i-1}
    \sim
    \mathcal{N}
    \left(
        \begin{bmatrix}
            m_{i|i-1}\\
            H_i m_{i|i-1}+\xi_i
        \end{bmatrix},
        \begin{bmatrix}
            P_{i|i-1}
            &
            P_{i|i-1}H_i^\top\\
            H_iP_{i|i-1}
            &
            S_i
        \end{bmatrix}
    \right).
    \label{eq:app_joint_predictive_distribution}
\end{equation}
Applying the conditional Gaussian formula yields
\begin{equation}
    s_i\mid\hat f_{0:i}
    \sim
    \mathcal{N}
    \left(
        m_{i|i},
        P_{i|i}
    \right),
    \label{eq:app_posterior_distribution}
\end{equation}
where
\begin{align}
    m_{i|i}
    &=
    m_{i|i-1}
    +
    K_i r_i,
    \label{eq:app_posterior_mean}\\
    P_{i|i}
    &=
    P_{i|i-1}
    -
    K_iS_iK_i^\top.
    \label{eq:app_posterior_covariance}
\end{align}
These are the recursive posterior-update equations used in Sec.~\ref{subsec:recursive_posterior_estimation}.

\subsection{LMMSE Optimality under Non-Gaussian Noise}
\label{subsec:lmmse_optimality}

This subsection proves that our method remains optimal in the LMMSE sense even when the process and observation noises in the model are not Gaussian. We only assume that the process and observation noises are zero mean, have finite second moments
\begin{equation}
Q_i=\operatorname{Cov}(\eta_i),
\qquad
R_i=\operatorname{Cov}(\varepsilon_i),
\label{eq:app_lmmse_noise_covariances}
\end{equation}
and are mutually and temporally uncorrelated and uncorrelated with the initial state. Recall the state-space model
\begin{equation}
\begin{aligned}
s_i &= A_i s_{i-1}+b_i+G\eta_i,\\
\hat f_i &= H_i s_i+\xi_i+\varepsilon_i.
\end{aligned}
\label{eq:app_lmmse_state_space}
\end{equation}
Assume that $m_{i-1|i-1}$ is already the LMMSE estimator of $s_{i-1}$ based on the observation history $\hat f_{0:i-1}$, with estimation error 
\begin{equation}
e_{i-1|i-1}
=
s_{i-1}-m_{i-1|i-1}
\end{equation}
and its covariance
\begin{equation}
P_{i-1|i-1}
=
\operatorname{Cov}\left(
e_{i-1|i-1}
\right).
\end{equation}
By the orthogonality principle, the estimation error $e_{i-1|i-1}$ has zero mean and is uncorrelated with $\hat f_{0:i-1}$. Define the prediction as
\begin{equation}
m_{i|i-1}=A_i m_{i-1|i-1}+b_i,
\label{eq:app_lmmse_prediction_mean}
\end{equation}
and define the prediction error and innovation by
\begin{equation}
e_{i|i-1}=s_i-m_{i|i-1},
\qquad
r_i=\hat f_i-\left(H_i m_{i|i-1}+\xi_i\right).
\end{equation}
Since
\begin{equation}
e_{i|i-1}=A_i e_{i-1|i-1}+G\eta_i,
\end{equation}
the assumptions imply that $e_{i|i-1}$ has zero mean and is uncorrelated with the past observations $\hat f_{0:i-1}$. Hence, $m_{i|i-1}$ is the LMMSE predictor of $s_i$ based on $\hat f_{0:i-1}$. Its error covariance is
\begin{equation}
P_{i|i-1}
=\operatorname{Cov}\left(e_{i|i-1}\right)
=A_iP_{i-1|i-1}A_i^\top+GQ_iG^\top.
\label{eq:app_lmmse_prediction_covariance}
\end{equation}
Additionally, using
\begin{equation}
r_i=H_i e_{i|i-1}+\varepsilon_i,
\end{equation}
it follows that $r_i$ also has zero mean and is uncorrelated with the past observations $\hat f_{0:i-1}$.
To construct the LMMSE estimator of $s_i$, consider the affine estimator
\begin{equation}
m_{i|i}=m_{i|i-1}+K_i r_i.
\label{eq:app_lmmse_affine_update}
\end{equation}
Its estimation error is
\begin{equation}
s_i-m_{i|i}=e_{i|i-1}-K_i r_i.
\end{equation}
This error has zero mean and is uncorrelated with the past observations $\hat f_{0:i-1}$. By the orthogonality principle, it remains to choose $K_i$ such that this error is uncorrelated with the innovation $r_i$, resulting in
\begin{equation}
\operatorname{Cov}\left(
e_{i|i-1}-K_i r_i,\,
r_i
\right)=0,
\end{equation}
or equivalently,
\begin{equation}
K_iS_i
=
\operatorname{Cov}\left(
e_{i|i-1},r_i
\right)
=
P_{i|i-1}H_i^\top,
\label{eq:app_lmmse_normal_equation}
\end{equation}
where
\begin{equation}
S_i
=
\operatorname{Cov}(r_i)
=
H_iP_{i|i-1}H_i^\top+R_i
\label{eq:app_lmmse_innovation_covariance}
\end{equation}
denotes the innovation covariance.
Thus, the optimal gain is
\begin{equation}
K_i
=
P_{i|i-1}H_i^\top S_i^{-1},
\label{eq:app_lmmse_kalman_gain}
\end{equation}
which is exactly the Kalman gain used in the recursive update. 
The corresponding estimation error covariance is
\begin{align}
P_{i|i}
&=
\operatorname{Cov}\left(
e_{i|i-1}-K_i r_i
\right) \notag\\
&=
P_{i|i-1}
-K_iH_iP_{i|i-1}
-P_{i|i-1}H_i^\top K_i^\top
+K_iS_iK_i^\top \notag\\
&=
P_{i|i-1}-K_iS_iK_i^\top.
\label{eq:app_lmmse_error_covariance}
\end{align}
Consequently, given the previous LMMSE estimate, the Kalman update is the LMMSE estimator based on the observation history $\hat f_{0:i}$.
Applying this argument recursively over the denoising steps establishes the LMMSE optimality of the complete recursion.

\section{Calibration of Statistics}
\label{app:calibration_statistics}

This appendix describes the calibration granularity and the estimation of the observation and process statistics used by QuAKE.

\paragraph{Granularity.}
Following the channel-wise calibration strategy commonly used in quantized diffusion correction methods~\citep{he2023ptqd,yao2024timestep,liu2025error,ryu2026q}, we estimate all statistics separately for each sampling step and each latent channel. For each timestep-channel pair, all spatial positions are aggregated to estimate one set of calibrated statistics. This aggregation reduces the amount of calibration data required for stable estimation. It corresponds to assuming that elements at the same timestep and channel share the same prior, observation, and process-noise parameters. During sampling, these calibrated statistics are broadcast to all spatial positions, while the correction itself is applied independently to each element. This element-wise correction avoids estimating cross-channel or cross-spatial covariance terms and keeps the online filtering cost small. For a fair comparison, the previous sampling-stage methods use the same timestep-wise and channel-wise calibration granularity.

\paragraph{Observation statistics.}
We collect calibration statistics along full-precision sampling trajectories. At each calibration step, we run both the full-precision denoiser and the quantized denoiser on the same full-precision latent state, obtaining paired outputs $f_i$ and $\hat f_i$. For each timestep and channel, all expectations, variances, and covariances are computed over the calibration samples and spatial positions. The valid current-output entry of the initial prior $m_{0|-1}$ and $P_{0|-1}$ is initialized from the marginal mean and variance of the first full-precision denoiser output. For later observation updates, QuAKE uses the conditional Gaussian model
\begin{equation}
    \hat f_i \mid f_i
    \sim
    \mathcal{N}(\gamma_i f_i+\xi_i,R_i).
\end{equation}
The observation gain and bias are estimated by linear regression from $f_i$ to $\hat f_i$:
\begin{equation}
    \gamma_i
    =
    \frac{\operatorname{Cov}(f_i,\hat f_i)}
    {\operatorname{Var}(f_i)},
    \qquad
    \xi_i
    =
    \mathbb{E}[\hat f_i]
    -
    \gamma_i\mathbb{E}[f_i].
\end{equation}
The residual variance is estimated as
\begin{equation}
    R_i
    =
    \operatorname{Var}
    \left(
        \hat f_i-\gamma_i f_i-\xi_i
    \right).
\end{equation}

\paragraph{Process statistics.}
The process statistics are calibrated on the full-precision sampling trajectories. For each step with available history, we apply the transition predictor used by QuAKE to the historical full-precision denoiser outputs and compare the prediction with the actual full-precision output at the current step. The difference is treated as the process residual $w_i$. For each timestep and channel, we estimate
\begin{equation}
    \mu_i
    =
    \mathbb E[w_i],
    \qquad
    Q_i
    =
    \operatorname{Var}(w_i).
\end{equation}
These statistics provide the process bias and process-noise variance used in the Kalman prediction step.

\begin{table*}[t]
\centering
\caption{
Quantitative results on MJHQ-30K and sDCI using the UniPC sampler. KID values are reported after multiplication by $10^3$. IR stands for ImageReward. Boldface marks the best result among sampling-stage correction methods for each model.
}
\label{tab:results_unipc}
\scriptsize
\setlength{\tabcolsep}{3.5pt}
\renewcommand{\arraystretch}{1.18}
\begin{tabular}{cc*{10}{c}}
\toprule
\multirow{3}{*}{Model} & \multirow{3}{*}{Method}
& \multicolumn{5}{c}{MJHQ-30K} & \multicolumn{5}{c}{sDCI} \\
\cmidrule(lr){3-7}\cmidrule(lr){8-12}
& & \multicolumn{2}{c}{Similarity ($\downarrow$)} & \multicolumn{3}{c}{Quality ($\uparrow$)}
& \multicolumn{2}{c}{Similarity ($\downarrow$)} & \multicolumn{3}{c}{Quality ($\uparrow$)} \\
\cmidrule(lr){3-4}\cmidrule(lr){5-7}\cmidrule(lr){8-9}\cmidrule(lr){10-12}
& &
FID & KID & CLIPScore & CLIP-IQA & IR
& FID & KID & CLIPScore & CLIP-IQA & IR \\
\midrule
\multirow{6}{*}{\shortstack{FLUX.1-dev}} & Full-Precision & -- & -- & 0.3186 & 0.9019 & 0.9029 & -- & -- & 0.3123 & 0.8473 & 0.8539 \\
 & Quantized & 7.45 & 0.264 & 0.3172 & 0.9078 & 0.8874 & 7.38 & 0.316 & 0.3115 & 0.8544 & 0.8541 \\
\cmidrule(lr){2-12}
 & Bias Correction & 7.35 & 0.294 & 0.3172 & 0.9070 & 0.8843 & 7.34 & 0.346 & 0.3118 & 0.8529 & 0.8439 \\
 & TAC-Diffusion & 7.39 & 0.285 & 0.3172 & \textbf{0.9081} & 0.8848 & 7.47 & 0.358 & \textbf{0.3126} & \textbf{0.8555} & \textbf{0.8527} \\
 & Q-Drift & 7.17 & 0.197 & \textbf{0.3175} & 0.9077 & \textbf{0.8853} & 7.14 & \textbf{0.168} & 0.3112 & 0.8542 & 0.8473 \\
\rowcolor{oursgray}
\cellcolor{white} & QuAKE (Ours) & \textbf{7.05} & \textbf{0.130} & 0.3173 & 0.9068 & 0.8797 & \textbf{7.06} & 0.180 & 0.3120 & 0.8548 & 0.8452 \\
\midrule
\multirow{6}{*}{\shortstack{Sana-0.6B}} & Full-Precision & -- & -- & 0.3372 & 0.9019 & 1.0923 & -- & -- & 0.3296 & 0.9145 & 0.9042 \\
 & Quantized & 10.63 & 1.575 & 0.3355 & 0.8950 & 1.0459 & 10.70 & 1.636 & 0.3275 & 0.9041 & 0.8604 \\
\cmidrule(lr){2-12}
 & Bias Correction & 23.10 & 8.128 & 0.3319 & 0.8761 & 0.9533 & 23.60 & 9.313 & 0.3228 & 0.8724 & 0.7448 \\
 & TAC-Diffusion & 22.72 & 7.954 & 0.3318 & 0.8773 & 0.9577 & 23.39 & 9.051 & 0.3223 & 0.8703 & 0.7385 \\
 & Q-Drift & 9.66 & 1.033 & 0.3357 & 0.8897 & 1.0351 & 10.00 & 1.283 & 0.3281 & 0.9018 & 0.8350 \\
\rowcolor{oursgray}
\cellcolor{white} & QuAKE (Ours) & \textbf{8.12} & \textbf{0.364} & \textbf{0.3363} & \textbf{0.8964} & \textbf{1.0523} & \textbf{8.31} & \textbf{0.409} & \textbf{0.3287} & \textbf{0.9072} & \textbf{0.8419} \\
\midrule
\multirow{6}{*}{\shortstack{PixArt-$\alpha$}} & Full-Precision & -- & -- & 0.3301 & 0.8905 & 0.9962 & -- & -- & 0.3141 & 0.9141 & 0.5903 \\
 & Quantized & 16.26 & 3.273 & 0.3279 & 0.8956 & 0.9145 & 16.07 & 3.739 & 0.3123 & 0.9127 & 0.4917 \\
\cmidrule(lr){2-12}
 & Bias Correction & 29.44 & 10.088 & 0.3279 & \textbf{0.9029} & 0.8977 & 30.87 & 12.498 & 0.3154 & \textbf{0.9214} & 0.4491 \\
 & TAC-Diffusion & 29.09 & 10.016 & 0.3280 & 0.9008 & \textbf{0.9037} & 30.31 & 12.323 & \textbf{0.3156} & 0.9145 & 0.4455 \\
 & Q-Drift & 13.77 & 2.183 & 0.3280 & 0.8851 & 0.8896 & 13.66 & 2.493 & 0.3114 & 0.9020 & 0.4583 \\
\rowcolor{oursgray}
\cellcolor{white} & QuAKE (Ours) & \textbf{13.38} & \textbf{1.918} & \textbf{0.3282} & 0.8952 & 0.9026 & \textbf{13.49} & \textbf{2.337} & 0.3126 & 0.9162 & \textbf{0.4885} \\
\midrule
\multirow{6}{*}{\shortstack{PixArt-$\Sigma$}} & Full-Precision & -- & -- & 0.3318 & 0.9065 & 0.9571 & -- & -- & 0.3238 & 0.9151 & 0.8587 \\
 & Quantized & 17.65 & 4.252 & 0.3299 & 0.8988 & 0.8669 & 17.22 & 4.831 & 0.3289 & 0.8883 & 0.7970 \\
\cmidrule(lr){2-12}
 & Bias Correction & 25.08 & 7.769 & 0.3294 & 0.8872 & 0.8746 & 24.55 & 9.013 & 0.3293 & 0.8716 & 0.7771 \\
 & TAC-Diffusion & 24.80 & 7.622 & 0.3300 & 0.8900 & 0.8628 & 24.22 & 8.788 & \textbf{0.3313} & 0.8763 & 0.7726 \\
 & Q-Drift & 17.21 & 4.073 & 0.3296 & 0.8939 & 0.8518 & 16.69 & 4.459 & 0.3293 & 0.8781 & 0.7895 \\
\rowcolor{oursgray}
\cellcolor{white} & QuAKE (Ours) & \textbf{15.82} & \textbf{3.270} & \textbf{0.3303} & \textbf{0.9021} & \textbf{0.8868} & \textbf{15.44} & \textbf{3.859} & 0.3291 & \textbf{0.8934} & \textbf{0.8030} \\
\bottomrule
\end{tabular}
\end{table*}

\begin{figure*}[t]
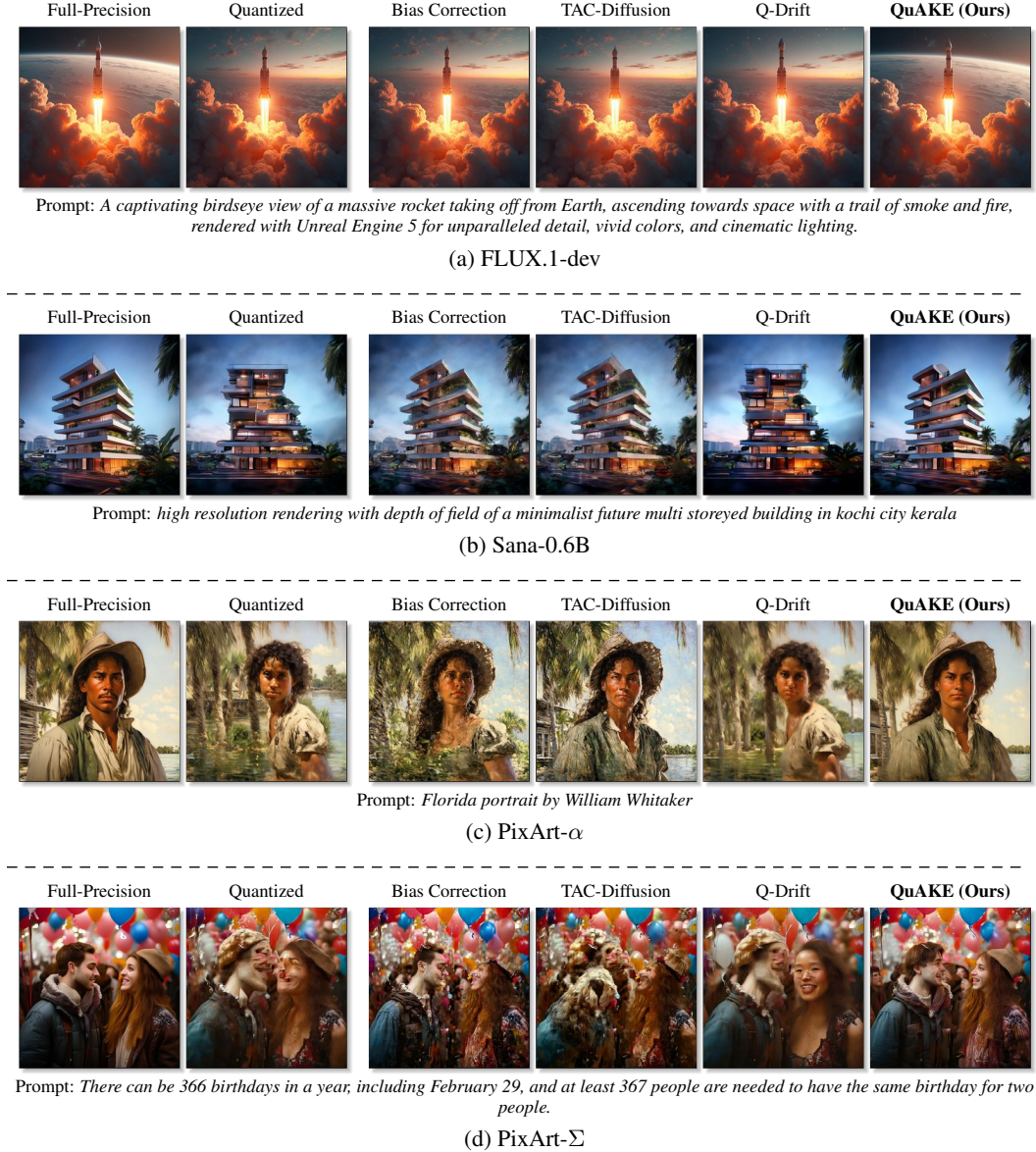

    \centering

    \mainresultrow
    {appendix}
    {flux}
    {(a) FLUX.1-dev}
    {A captivating birdseye view of a massive rocket taking off from Earth, ascending towards space with a trail of smoke and fire, rendered with Unreal Engine 5 for unparalleled detail, vivid colors, and cinematic lighting.}

    \hdashrule{\linewidth}{0.5pt}{4pt 3pt}

    \mainresultrow
    {appendix}
    {sana}
    {(b) Sana-0.6B}
    {high resolution rendering with depth of field of a minimalist future multi storeyed building in kochi city kerala}

    \hdashrule{\linewidth}{0.5pt}{4pt 3pt}

    \mainresultrow
    {appendix}
    {alpha}
    {(c) PixArt-$\alpha$}
    {Florida portrait by William Whitaker}

    \hdashrule{\linewidth}{0.5pt}{4pt 3pt}

    \mainresultrow
    {appendix}
    {sigma}
    {(d) PixArt-$\Sigma$}
    {There can be 366 birthdays in a year, including February 29, and at least 367 people are needed to have the same birthday for two people.}

    \caption{Additional qualitative comparison across different text-to-image diffusion models using UniPC.}
    \label{fig:appendix_results}
\end{figure*}

\section{Implementation Details}
\label{app:implementation_details}
This appendix provides additional implementation details for QuAKE, focusing on how the linear transition is constructed and how the output window is handled at the beginning of the sampling trajectory.

\paragraph{Transition coefficients.}
We construct the linear Gaussian transition through an $x_0$-domain extrapolation. Empirically, denoiser trajectories are smoother after converting model outputs to the predicted $x_0$ parameterization and using the log-SNR coordinate $\lambda$ as the extrapolation variable~\citep{lu2022dpm,lu2025dpm}. This leads to smaller process residuals in practice for QuAKE. Therefore, at each step, we first convert the available historical model outputs to the corresponding predicted $x_0$ values, perform Lagrange extrapolation in the $\lambda$ coordinate, and then convert the extrapolated $x_0$ back to the model-output parameterization required by the sampler. This construction remains compatible with the linear Gaussian state-space model used by our method. For fixed sample states and noise levels, the conversion between the model-output parameterization and the predicted $x_0$ parameterization is affine. Since Lagrange extrapolation is linear in the available historical values, the whole procedure is an affine function of the previous model outputs. Therefore, it can be absorbed into the transition matrix and bias term, and the process model still takes the form
\begin{equation}
    s_i
    =
    A_i s_{i-1}
    +
    b_i
    +
    G\eta_i,
\end{equation}
where the first row of $A_i$ and the corresponding entry of $b_i$ are induced by the composed mapping from model outputs to the $x_0$-domain, the $\lambda$-domain Lagrange extrapolation, and the mapping from the extrapolated $x_0$ back to the model-output domain. The remaining rows of $A_i$ implement the standard history-shift operation, so that the corrected current output becomes a historical entry in the next step.

\paragraph{Boundary handling.}
At the first sampling step, there is no previous denoiser output to propagate. We therefore skip the transition prediction and directly apply the observation update to the initial prior. The posterior state after this step contains only the corrected current output, and historical entries that do not correspond to valid denoiser outputs are not returned to the sampler. During the first few sampling steps, the number of available historical outputs can be smaller than the state window length. In this case, QuAKE uses the same $x_0$-domain extrapolation procedure with only the available history. Equivalently, the transition is formed with a reduced effective order, while the state is kept in the same augmented format for implementation consistency. After each update, the sampler reads only the valid posterior entries required at the current step. Once enough historical outputs are available, QuAKE uses the full-order Lagrange extrapolation in the $\lambda$ coordinate.

\section{Detailed Experimental Setup}
\label{app:detailed_experimental_setup}

This appendix provides additional details on the model configurations, quantization settings, evaluation metrics, and implementation environment used in our experiments. Unless otherwise specified, all models are evaluated with their default inference configurations. For text-to-image sampling, we use a resolution of $1024\times1024$ and a classifier-free guidance scale of $4.5$ for PixArt-$\alpha$, PixArt-$\Sigma$, and Sana-0.6B, while FLUX.1-dev is evaluated at a resolution of $512\times512$ with a guidance scale of $3.5$. For quantized models, we only quantize the main diffusion backbone, while keeping the VAE, text encoder, and other auxiliary components in their original precision. For the SVDQuant model-quantization hyperparameter $r$, we set $r=32$ for PixArt-$\alpha$ and PixArt-$\Sigma$, $r=16$ for FLUX.1-dev, and $r=4$ for Sana-0.6B. For evaluation, FID and KID are computed using \texttt{clean-fid}~\citep{parmar2022aliased}. CLIP-based evaluation metrics are computed with the CLIP-ViT-B/32 backbone~\citep{radford2021learning}. All experiments are conducted on NVIDIA A100-SXM4-40GB GPUs.

\section{Additional Results}
\label{app:additional_results}

This appendix provides additional quantitative and qualitative results to further evaluate the generalization of QuAKE across different numerical samplers. Table~\ref{tab:results_unipc} reports results using the UniPC sampler. QuAKE consistently provides the strongest overall distributional recovery across different models and datasets, substantially reducing the FID and KID gap to the full-precision generation while maintaining competitive generation quality. These results closely follow the trend observed with DPM-Solver++, demonstrating that the effectiveness of QuAKE is not tied to a specific numerical solver and further supporting its applicability to different high-order multistep ODE samplers.

Figure~\ref{fig:appendix_results} provides additional qualitative comparisons. Consistent with the quantitative results, QuAKE effectively compensates for quantization-induced generation deviations across different model architectures and produces outputs that more closely resemble the corresponding full-precision generations than existing sampling-stage correction methods.

\end{document}